\documentclass{article} %
\usepackage{iclr2023_conference,times}

\usepackage{amsmath,amsfonts,bm}
 \usepackage{booktabs}

\def\eqref#1{equation~\ref{#1}}

\def\1{\bm{1}}

\def\eps{{\varepsilon}}

\DeclareMathAlphabet{\mathsfit}{\encodingdefault}{\sfdefault}{m}{sl}
\SetMathAlphabet{\mathsfit}{bold}{\encodingdefault}{\sfdefault}{bx}{n}

\usepackage{hyperref}
\usepackage{url}
\usepackage{graphicx}
\usepackage{subfig}
\usepackage{dirtytalk}
\usepackage{amsthm}
\usepackage{amsmath}
\usepackage{soul}
\usepackage{amssymb}
\usepackage{xspace}
\usepackage{mathtools}

\usepackage{algorithm}
\usepackage{algpseudocode}
\algnewcommand\algorithmicinput{\textbf{Input:}}
\algnewcommand\algorithmicoutput{\textbf{Output:}}
\algnewcommand\Input{\item[\algorithmicinput]}%
\algnewcommand\Output{\item[\algorithmicoutput]}%

\newtheorem{definition}{Definition}

\newcommand{\ip}[2]{\langle #1, #2 \rangle}

\newcommand{\gaussian}{\mathcal{N}}
\newcommand{\can}{z} %
\newcommand{\canarygrad}{\nabla_\theta \ell(\can)}
\newcommand{\prob}{\mathbb{P}}

\author{Samuel Maddock\thanks{Work done during an internship at Meta. 
}\\
University of Warwick\\
\And
Alexandre Sablayrolles \\
Meta AI \\
\And
Pierre Stock \\
Meta AI \\
}

\newcommand{\edit}[1]{{#1}}
\newcommand{\fedit}[1]{{#1}}

\def\method{\textsc{Canife}\@\xspace}
\iclrfinalcopy %
\begin{document}

\setcounter{page}{1}

\title{\method: Crafting Canaries for Empirical Privacy Measurement in Federated Learning}
\maketitle

\begin{abstract}
Federated Learning (FL) is a setting for training machine learning models in distributed environments where the clients do not share their raw data but instead send model updates to a server. However, model updates can be subject to attacks and leak private information. Differential Privacy (DP) is a leading mitigation strategy which involves adding noise to clipped model updates, trading off performance for strong theoretical privacy guarantees. Previous work has shown that the threat model of DP is conservative and that the obtained guarantees may be vacuous or may overestimate information leakage in practice. In this paper, we aim to achieve a tighter measurement of the model exposure by considering a realistic threat model. We propose a novel method, \method, that uses \emph{canaries}---carefully crafted samples by a strong adversary to evaluate the empirical privacy of a training round. We apply this attack to vision models trained on CIFAR-10 and CelebA and to language models trained on Sent140 and Shakespeare. 
In particular, in realistic FL scenarios, we demonstrate that the empirical per-round epsilon obtained with \method is \fedit{4 -- 5$\times$} lower than the theoretical bound. 
\end{abstract}

\section{Introduction}\label{sec:intro}
\looseness=-1 Federated Learning (FL) has recently become a popular paradigm for training machine learning models across a large number of clients, each holding local data samples \citep{mcmahan2017communication}. 
The primary driver of FL's adoption by the industry is its compatibility with the \say{privacy by design} principle, since the clients' raw data are not communicated to other parties during the training procedure~\citep{kairouz2019advances,huba2022papaya,xu2022training}.
Instead, clients train the global model locally before sending back updates, which are aggregated by a central server.
However, model updates, in their individual or aggregate form, leak information about the client local samples~\citep{geiping2020inverting,gupta2022recovering}. 

Differential Privacy (DP) \citep{dwork2006calibrating,abadi2016deep} is a standard mitigation to such privacy leakage. Its adaptation to the FL setting, %
\textsc{DP-FedAvg}~\citep{mcmahan2017learning}, provides user-level guarantees by adding Gaussian noise to the aggregated clipped model updates received by the server. In practice, training with strong privacy guarantees comes at the expense of model utility~\citep{bassily2014differentially,kairouz2019advances}, notwithstanding  efforts to close this gap, either with public pre-training and partial model updates~\citep{xu2022training}, accountants with better compositionality properties~\citep{mironov2017renyi} or DP variants such as \textsc{DP-FTRL}~\citep{kairouz2021practical}.

Hence, it is common in practical deployments of DP-FL to train with a high privacy budget $\varepsilon$ resulting in loose privacy guarantees~\citep{ramaswamy2020training}. Such large privacy budgets often provide vacuous guarantees on the information leakage, for instance, against membership inference attacks~\citep{mahloujifar2022optimal}. Encouragingly, recent work has shown that the information recovered in practice using state-of-the-art attacks is less than what theoretical bounds may allow~\citep{nasr2021adversary}. This suggests that DP is conservative and that a tighter measurement of the model exposure may be achieved by considering more realistic threat models. 

\begin{figure*}[t]
   \subfloat[\fedit{Sent140 (2-layer LSTM)}]{%
      \includegraphics[width=0.325\textwidth]{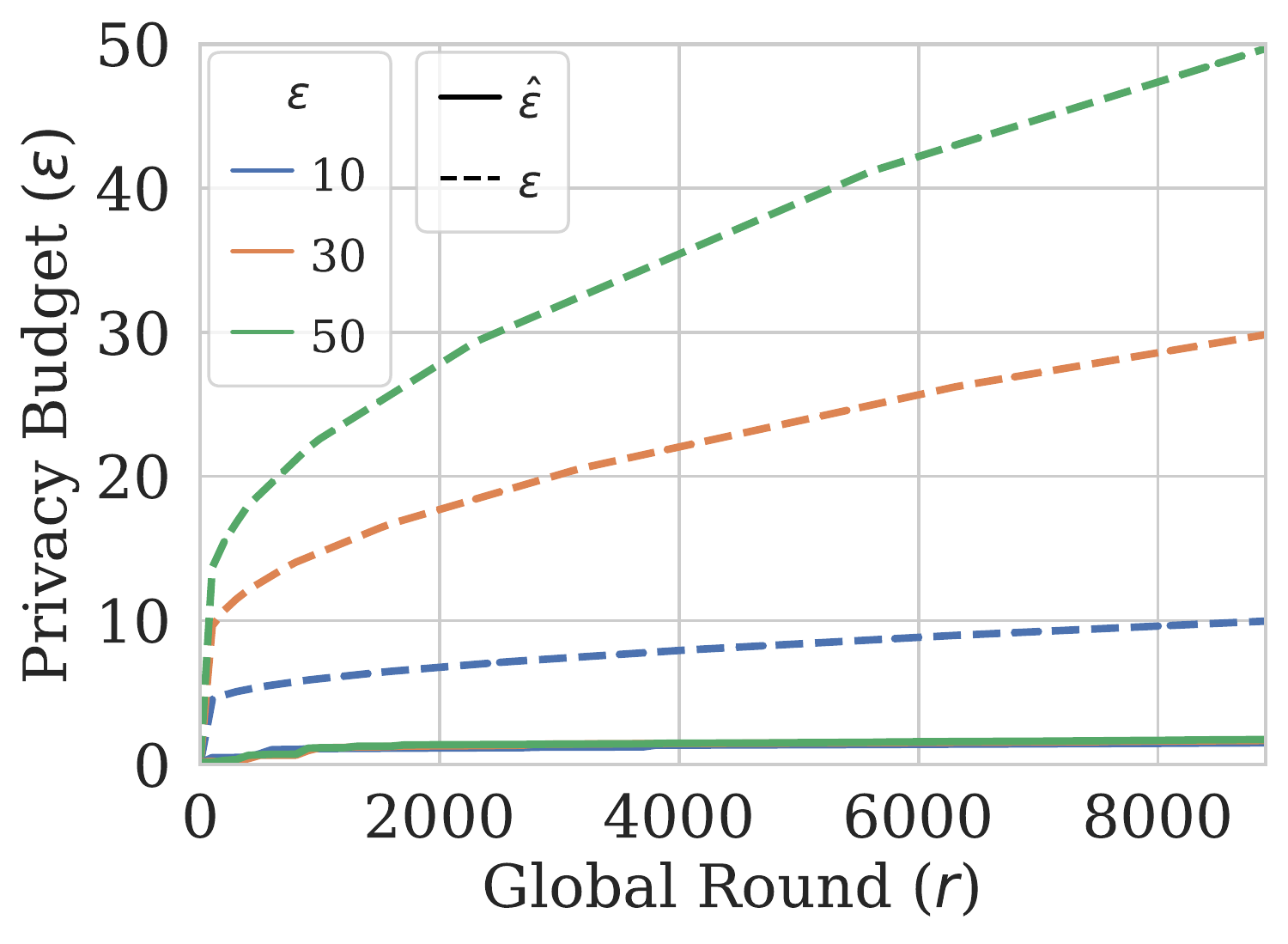}}
    \hspace{\fill}
   \subfloat[\fedit{CelebA (ResNet18)}]{%
      \includegraphics[width=0.33\textwidth]{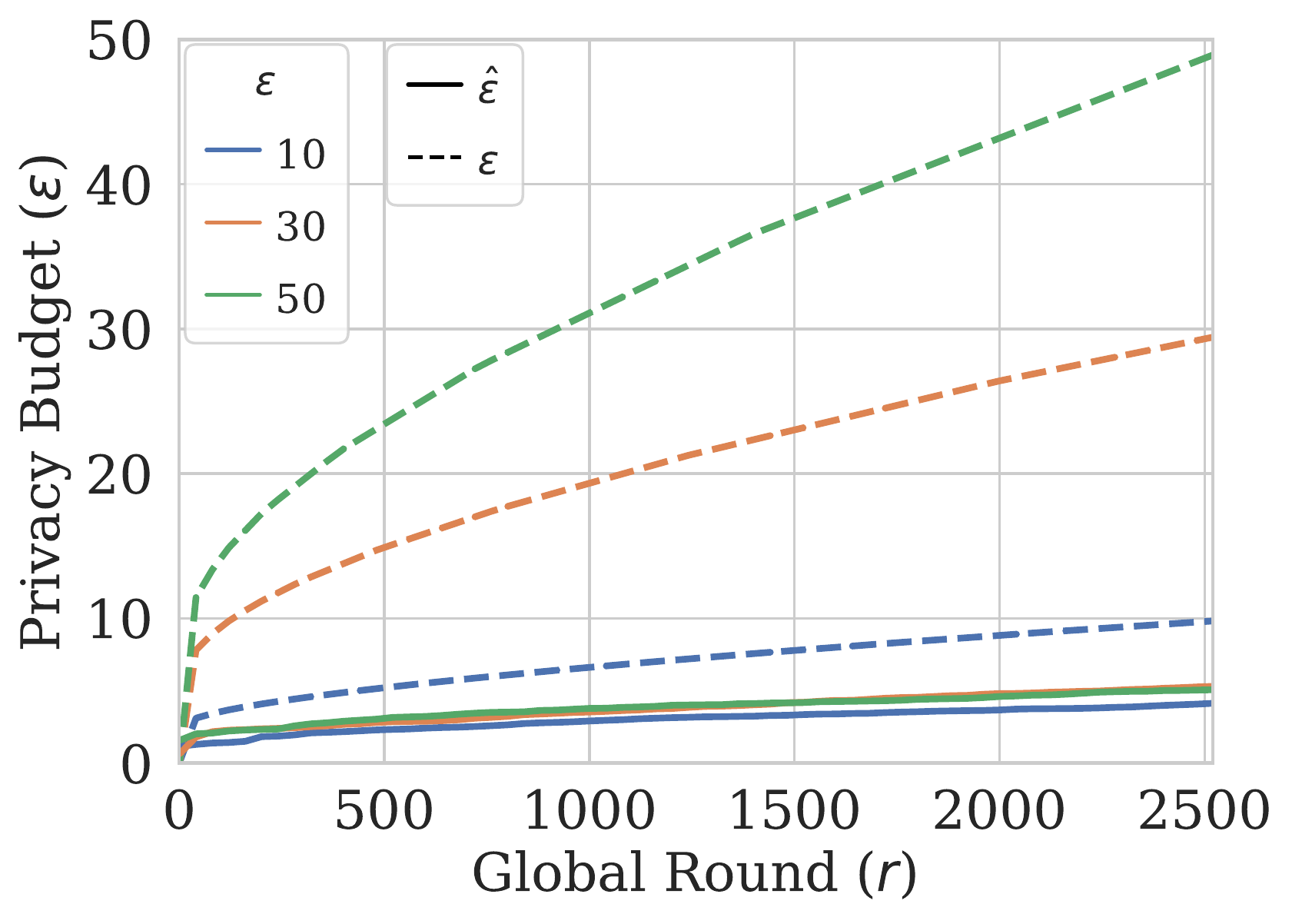}}
    \hspace{\fill}
   \subfloat[\fedit{Shakespeare (2-layer LSTM)}]{%
      \includegraphics[width=0.33\textwidth]{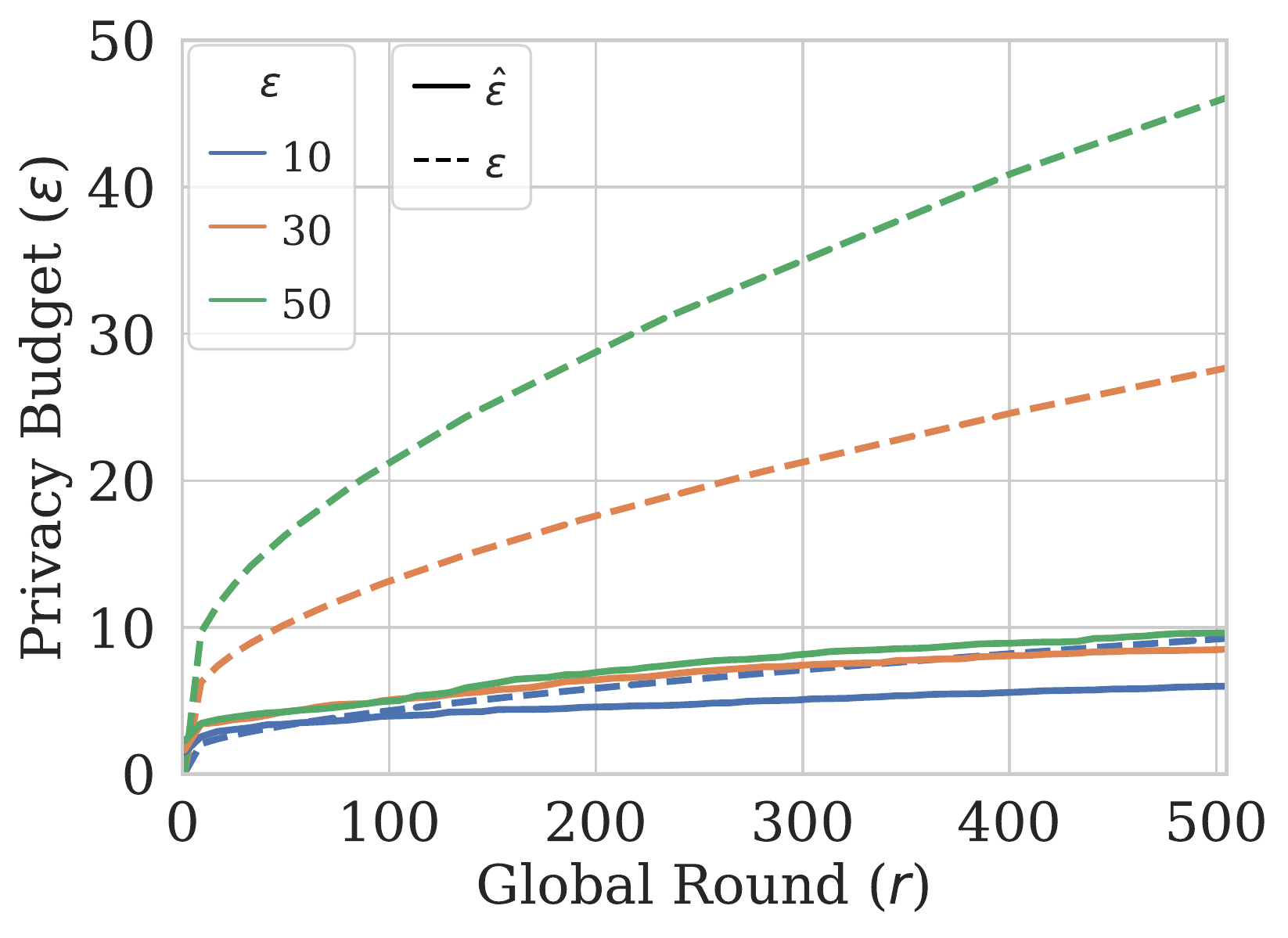}}
    \\
\vspace{-3mm}
\caption{Empirical privacy measurements over the course of FL training for %
LEAF benchmarks Sent140, CelebA and Shakespeare with $\eps \in \{10,30,50\}$. We observe a notable gap between the theoretical $\eps$ obtained with \textsc{DP-FedSGD} and the empirical $\hat \varepsilon$ obtained with \method.}
\label{fig:train_and_freeze}
\end{figure*}

\begin{figure}[t]
\begin{center}
\includegraphics[width=\textwidth, height=4cm]{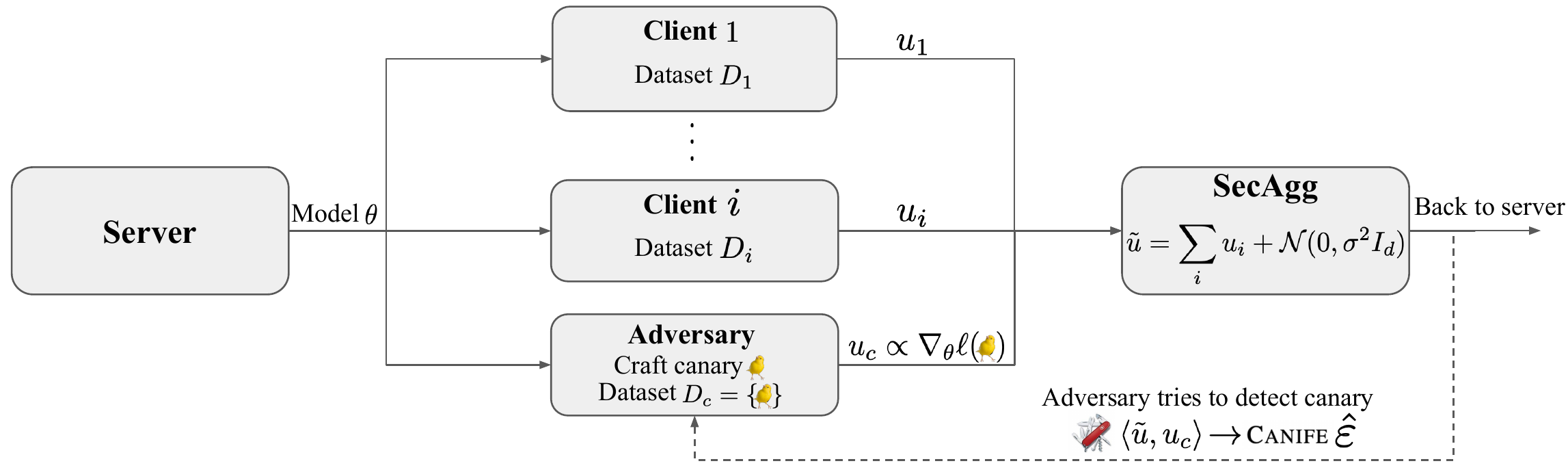}
\end{center}
\caption{Illustration of the proposed \method method for one training round. The adversary is a rogue client that crafts a training sample resulting in an extremely out-of-distribution model update $u_c$ and performs a membership inference attack on the (public) aggregated noisy model update $\tilde u$ provided to the server by SecAgg. Attack results are converted to empirical privacy guarantees $\hat \varepsilon$ for this round. We then compound these guarantees over the course over the training as in Section~\ref{sec:privacy_measure}.}
\label{fig:framework}
\end{figure}

In this paper, we propose to complement DP-FL training with a novel attack method, CANaries In Federated Environments (\method), to measure empirical privacy under a realistic threat model. 
We assume that a rogue client wants to reconstruct data samples from the model updates.
To make its job easier, this adversary is allowed to craft an outlier training sample, the \emph{canary}.
The training round proceeds normally, after which the rogue client performs a statistical test to detect the canary in the global noisy model update provided to the server by any secure aggregation protocol (see Figure~\ref{fig:framework}). 
Finally, we translate the attack results into a per round measure of empirical privacy~\citep{jagielski2020auditing, nasr2021adversary}  and propose a method using amplification by subsampling to compute the empirical privacy incurred during training as depicted in Figure~\ref{fig:train_and_freeze} for standard FL benchmarks.

Critically, our privacy attack is designed to approximate the worst-case \emph{data sample}, not the worst-case \emph{update vector}. The rogue client seeks to undermine the privacy guarantee by manipulating its input, which is consistent with FL environments using secure sandboxing to protect the integrity of the training process~\citep{frey2021introducing}. We additionally model the server as the honest party, not allowing it to poison the global model in order to reconstruct training samples, in contrast with a recent line of work~\citep{fowl2021robbing, boenisch2021curious,wen2022fishing,fowl2022decepticons}.

In summary, our contributions are as follows:
\begin{itemize}
    \item We propose \method (Section \ref{sec:method}), a novel and practical privacy attack on FL that injects crafted \emph{canary} samples. It augments the standard DP-FL training with a tight measure of the model's privacy exposure given a realistic yet conservative threat model.
    \item \method is compatible with natural language and image modalities, lightweight and requires little representative data and computation to be effective. As a sanity check, we demonstrate that \method tightly matches DP guarantees in a toy setup (Section \ref{sec:exp:cifar10}) before exploring how it behaves in the federated setting (Section \ref{sec:exp:monitor}). 
    \item Our work highlights the gap between the practical privacy leakage and the DP guarantees in various scenarios. \fedit{For instance, on the CelebA benchmark, we obtain an empirical measure $\hat \varepsilon \approx 6$ for a model trained with a formal privacy guarantee of $\varepsilon = 50$}. 
\end{itemize}

\section{Background}\label{sec:background}

\subsection{Differential Privacy}
Differential Privacy~\citep{dwork2006calibrating,dwork2014foundations} defines a standard notion of privacy that guarantees the output of an algorithm does not depend significantly on a single sample or user.

\begin{definition}[Differential Privacy]
A randomised algorithm $\mathcal{M}\colon \mathcal{D} \rightarrow \mathcal{R}$ satisfies $(\varepsilon, \delta)$-differential privacy if for any two adjacent datasets $D, D^\prime \in \mathcal{D}$ and any subset of outputs $S \subseteq \mathcal{R}$, $$\prob(\mathcal{M}(D) \in S) \leq e^\eps \prob(\mathcal{M}(D^\prime) \in S) + \delta.$$
\end{definition}

The privacy parameter $\eps$ is called the \emph{privacy budget} and it determines an upper bound on the information an adversary can obtain from the output of an $(\eps,\delta)$-DP algorithm. The parameter $\delta$ defines the probability of failing to guarantee the differential privacy bound for any two adjacent datasets. %
In this work, we are interested in \emph{user-level} differential privacy which takes $D$ and $D^\prime$ to be \emph{adjacent} if $D^\prime$ can be formed by adding or removing all samples associated with a single user from $D$. 

Standard DP results state that the privacy budget $(\varepsilon, \delta)$ accumulates (roughly) in proportion to the square root of the number of iterations. Advanced privacy accountants leverage the uncertainty due to random sampling~ \citep{mironov2017renyi,wang2019subsampled,gopi2021numerical}. We use the R\'enyi Differential Privacy (RDP) accounting implemented in the Opacus library \citep{yousefpour2021opacus}.

\subsection{Private Federated Learning}
A standard FL protocol, such as \textsc{FedAvg}~\citep{mcmahan2017communication} computes a weighted average of the model updates from clients before performing a gradient descent step on the global model. Other variants exist to deal with common optimization problems in the federated setting such as convergence speed \citep{wang2019matcha}, heterogeneity \citep{karimireddy2020mime, karimireddy2020scaffold}, reducing communication bandwidth \citep{alistarh2017qsgd} and adding momentum \citep{reddi2020adaptive}.
In this work, we focus on \textsc{DP-FedSGD}, a private extension of \textsc{FedAvg}~\citep{mcmahan2017learning}. At each round, the selected clients compute their clipped model update $u_i$ and transmit it to the server, which aggregates the model updates and adds Gaussian noise:
\[
    \tilde{u} = \sum_i u_i + \mathcal{N}(0, \sigma^2I_d).
\]
In practice, federated algorithms rely on Secure Aggregation (SegAgg) protocols to aggregate each client update without revealing any individual $u_i$ to the server or to other participants~\citep{bonawitz2017practical,bell2020secure}. 
In a TEE-based SecAgg~\citep{huba2022papaya}, a trusted execution environment (TEE) aggregates the individual model updates and calibrated DP noise before handing over the noisy model update $\tilde u$ to the server.
These specific models are orthogonal to our work and we assume from now on that the server and clients participate in a TEE-based SecAgg protocol. We discuss our threat model with regards to our attack in Section~\ref{sec:threat}.

\subsection{Attacks \& Empirical Privacy}

\paragraph{Centralized and FL Attacks.} There is a vast body of literature on attacking models trained in the centralised setting. %
For example, membership inference attacks (MIA) attempt to distinguish whether a sample was present in the training set given only the trained model \citep{shokri2017membership, sablayrolles2019white}. %
Others attacks consider the more difficult problem of reconstructing entire training samples from a trained model, often using (batch) gradient information~\citep{yin2021see, jeon2021gradient,balle2022reconstructing}. 
Since model updates $u_i$ are essentially just aggregated gradients, it is natural that FL updates may leak private information as well. \citet{nasr2019comprehensive} show that it is possible to perform both passive and active membership-inference attacks in the federated setting. Other works such as that of \citet{fowl2021robbing} and \citet{wen2022fishing} have designed attacks on model updates which allow for the reconstruction of training samples used in federated training. However, they assume that the server can poison the global model, whereas we assume an honest server.

\paragraph{Canaries.} The notion of canary samples usually refers to natural data samples used to measure memorization in large language models \citep{carlini2019secret,carlini2021extracting,thakkar2020understanding,shi2022just}. For instance, \cite{parikh2022canary} propose to reconstruct canaries inserted in training data and \cite{stock2022defending} insert canaries to track and quantify the information leakage when training a causal language model.
In all prior work, the canary is either a sample from the training set or a handcrafted instance such as ``My SSN is 123-45-6789''. In contrast, \method provides an explicit method for crafting canary samples that are as adversarial as possible within the given threat model
(see Appendix~\ref{appendix:canary_samples} for canary samples) to obtain tight measurement of the model's exposure. Moreover, the proposed method applies to any training modality allowing to backpropagate in the sample space (in particular, pixels for images and tokens for natural language).

\paragraph{Empirical Privacy.} The proposed approach departs from existing attacks in the FL setup.
For instance, \cite{jayaraman2019evaluating} and \cite{jagielski2020auditing} have derived empirical measures of privacy through attacks and often shown gaps between the empirically measured and theoretical privacy. More recently, \citet{nasr2021adversary} study a range of membership-inference attacks \fedit{by} varying adversary's powers.
We argue that the threat model of many of these attacks is often too permissive relative to what a realistic adversary can achieve in the federated setting. For example, attacks in \citet{nasr2021adversary} assume knowledge of the other samples in the dataset.

\section{Methodology}\label{sec:method}

The adversary wants to design a canary sample %
$\can$ to measure the empirical privacy of a given FL training round under a realistic threat model defined in Section~\ref{sec:threat}. 
We view this problem as a membership inference game where a rogue client crafts a canary $\can$ specific to the global model at round $r$ (Section~\ref{sec:canary_design}) and uses this canary to produce an extremely out-of-distribution model update. After this, the server proceeds as usual, aggregating model updates via any secure aggregation primitive and adding noise. At the end of the round, the adversary attempts to detect the presence of the canary in the aggregated noisy model update $\tilde{u}$ (Section~\ref{sec:canary_test}) and computes a per-round empirical privacy measure.
These per-round empirical privacy guarantees are compounded over the whole course of the training to give a final empirical privacy guarantee as explained in Section~\ref{sec:privacy_measure}.

\subsection{Threat Model}\label{sec:threat}

We assume an \emph{honest-but-curious} threat model where the clients and the server do not deviate from the training protocol~\citep{bonawitz2017practical,kairouz2019advances,huba2022papaya}. 
In particular, the clients cannot directly manipulate gradients nor the final model updates from the local training and the server cannot modify the model maliciously to eavesdrop on the clients. 

To craft the canary for a given training round, we assume that the adversary --- a rogue client --- has access to (1) the current server-side public model and to (2) emulated \emph{mock clients}. 
Mock clients are created using a public set of samples called the \emph{design pool} \edit{whose} distribution is similar to that of the clients.
We show in Section~\ref{sec:exp} that the adversary is able to design a strong and robust canary sample even with a small amount of public data and under mild assumptions about the true data distribution. 
We also demonstrate that the computational resources required to design the canary are small, which means that even adversaries under resource constraints are able to launch such attacks.

We argue such a threat model is realistic and discuss practical limitations that influence the adversary's success in Section~\ref{sec:rel2dp}: our goal is to design an attack that is as adversarial as possible (in order to derive a worst-case measure of empirical privacy) under a set of reasonable assumptions.

\subsection{Canary Detection}
\label{sec:canary_test}

\looseness=-1 The adversary's objective is to craft a canary $\can$ such that the resulting model update is extremely out-of-distribution. More precisely, we require that the canary gradient $\canarygrad$ is orthogonal to all individual clipped model updates $u_i$ in the current round: $\ip{u_i}{\canarygrad} = 0$. We demonstrate that this allows the adversary to detect the presence of the canary by separating two Gaussian distributions. 

The server trains a model parameterized by $\theta \in \mathbb R^d$ with a certain level of DP noise $\sigma$ across a population of clients each holding a local dataset $D_i$. Since the rogue client holds a dataset that only contains the canary $D_c = \{\can\}$, its clipped model update will be proportional\footnote{We assume that the client's local optimizer is SGD with no momentum \citep{mcmahan2017learning}.} to the canary gradient: $u_c \propto  \canarygrad$.
Recall that the aggregated private model update $\tilde{u}$ is formed by the noisy sum of individual clipped model updates: $\tilde{u} = \sum_i u_i + \mathcal{N}(0, \sigma^2 I_d)$. Then, as a counter-factual, if the rogue client does not participate in the training round:
\begin{align*}
    \left\langle \tilde{u}, u_c\right\rangle &\propto \sum_i \underbrace{\vphantom{\left\langle\mathcal{N}(0, \sigma^2 I_d), \canarygrad\right\rangle}\left\langle u_i, \canarygrad\right\rangle}_{\vphantom{\mathcal N(0, \sigma^2\|\canarygrad\|^2)}=0 \text{ by design}} + \underbrace{\left\langle\mathcal{N}(0, \sigma^2 I_d), \canarygrad\right\rangle}_{=\mathcal N(0, \sigma^2\|\canarygrad\|^2)}.
\end{align*}
Hence, $\left\langle \tilde{u}, u_c\right\rangle$ follows a %
one-dimensional zero-mean Gaussian with variance $\sigma_c^2$, where $\sigma_c$ accounts for the proportionality factor. Similarly, if the rogue client participates in the training round, 
$\left\langle \tilde{u}, u_c\right\rangle$ follows a one-dimensional Gaussian with the same variance $\sigma_c^2$ centered at $\|u_c\|^2$. Thus, %
the membership inference game is reduced to separating two Gaussian distributions centered at $0$ and \edit{$||u_c||^2$ respectively}. %
Note that the result is unchanged up to a fixed scaling factor if the client updates~$u_i$ are weighted by their local dataset size as in~\citep{mcmahan2017learning}.
The testing approach is described in full in Algorithm~\ref{alg:design} and involves computing the attack score $s_r := \left\langle \tilde{u}, u_c\right\rangle$.  %
We derive a connection with the likelihood ratio test in Appendix~\ref{appendix:lkelihood}.

\subsection{Canary Design}
\label{sec:canary_design}
At a given round $r$, the rogue client is given the current server-side model parameterized by $\theta$. Given a set of heldout clipped model updates $\{u_i\}$,  it creates the canary by minimizing:
\begin{align}\label{eq:loss2}
    \mathcal L(\can) = \sum_i \ip{u_i}{\canarygrad}^2 + \max(C-||\canarygrad||, 0)^2.
\end{align}
Recall that $\canarygrad$ denotes the gradient of the network's parameters with respect to its training loss $\ell$ when forwarding the canary $\can$ through the network. 
The first loss term is designed to make the canary gradient $\canarygrad$ orthogonal to the set of heldout model updates while the second term enforces the canary norm is not smaller than some constant, that we set to the clipping constant $C$ of DP-FL. \edit{This ensures $||u_c||^2 = C^2$ and for simplicity we fix $C=1$.} 
In Appendix \ref{appendix:ablation}, we provide experiments that show that choosing the gradient norm constant too large \edit{(i.e., much larger than $C$)} has a detrimental effect on optimization.

Using an automatic differentiation framework such as PyTorch~\citep{paske2019pytorch}, we compute~$\nabla_z \mathcal L(z)$ and perform stochastic gradient descent directly in the sample space as described in Algorithm~\ref{alg:design}. (Recall that the model parameters $\theta$ are fixed during the optimization procedure.) Computing $\nabla_z \mathcal L(z)$ is straightforward for continuous data, such as images, as we can simply backpropagate in the pixel space. 
For language models, the problem is more complex as the input is a sequence of discrete tokens. 
Hence, we leverage the work of \citet{guo2021gradientbased}, who use the Gumbel-Softmax distribution \citep{jang2016categorical} to forge adversarial language examples. 
This allows them to use gradient-based methods by optimising a probability distribution over each token in a sequence.

We investigate various methods to initialize the canary, including starting from a random training sample or random pixels or tokens. Depending on the task at hand, we might need to fix a target for the canary. For instance, for image classification, we need to assign the canary to a class in order to be able to compute $\canarygrad$. We investigate various canary initialization and target choice strategies experimentally in Section~\ref{sec:exp}. We also investigate other optimization considerations for designing the canary such as slightly modifying the loss in Appendix~\ref{appendix:ablation}.

\paragraph{Adversarial Examples.}
We can view the canary $\can$ as \say{adversarial} in the sense that it should be extremely out-of-distribution to get a worst-case measure of privacy. This %
is different from the unrelated notion of \emph{adversarial examples} which typically constrain the changes of a sample to be imperceptible to humans \citep{biggio2013evasion}. In our setup, we do not impose this constraint as we wish to encompass the realistic worst-case of an extreme out-of-distribution sample.

\begin{algorithm}[t]
\caption{\method attack by a rogue client} \label{alg:design}
\begin{algorithmic}[1] 
\Input Design pool $D_{\text{pool}}$, Design iterations $T$, Canary learning rate $\beta$, Global model $\theta$ 
    \State Form mock clients from the design pool $D_{\text{pool}}$
    \State For each mock client $i$, compute the clipped model update $u_i$
    \State Initialise the canary $\can_0$ \Comment{See Section~\ref{sec:canary_design}}
    \For{$t=1, \dots, T$}
        \State $\mathcal L(\can_t) \leftarrow \sum_i \ip{u_i}{C \cdot \nabla_\theta \ell(\can_t)}^2 + \max(C-||\nabla_\theta \ell(\can_t)||, 0)^2$ \Comment{ Canary optimization loss}
        \State Compute $\nabla_{\can_t}\mathcal{L}(\can_t)$ \Comment{Gradient of the canary loss w.r.t $\can_t$}
        \State $\can_{t+1} \leftarrow \can_t - \beta \cdot \nabla_{\can_t}\mathcal{L}(\can_t)$  \Comment{For NLP, see Section~\ref{sec:canary_design}}
    \EndFor
    \State Compute $u_c$  \Comment{Model update with $D_c = \{(\can_T, y_c)\}$} 
    \State \Return $s_r \leftarrow \langle \tilde{u}, u_c \rangle$ \Comment{$\tilde u$ is deemed public after the round has finished}
\end{algorithmic}
\end{algorithm}

\subsection{Measuring Empirical Privacy}
\label{sec:privacy_measure}

The \method attack is carried out over a certain number ($n$) of fake rounds where the server does not update its global model but where the pool of selected (regular) clients differs every time. Hence, \method has no influence on the model's final accuracy. The adversary crafts a single canary and inserts it into every fake round with probability $1/2$. Once the $n$ attack scores are computed, the adversary deduces the performance of the attack at a calibrated threshold $\gamma$. 
Building on previous work~\citep{jagielski2020auditing, nasr2021adversary}, we compute the empirical privacy guarantee  $\hat{\varepsilon}$ based on the False Positive Rate (FPR) and False Negative Rate (FNR) of the attack at the threshold $\gamma$ as
\begin{align*}
  \hat{\eps} = \max\left(\log \frac{1-\delta-\mathrm{FPR}}{\mathrm{FNR}},\log \frac{1-\delta-\mathrm{FNR}}{\mathrm{FPR}}\right).
\end{align*}
Our measure differs slightly from that of \citep{nasr2021adversary} as our attack measures privacy at a single round of training. Thus, $\hat{\varepsilon}$ is a \emph{per-round} privacy measure which we denote as $\hat{\eps}_r$ and we compare this to a per-round theoretical $\eps_r$. 
In order to deduce an overall empirical epsilon $\hat{\eps}$ we convert the per-round $\hat{\eps}_r$ into a noise estimate $\hat{\sigma}_r$ under amplification by subsampling. We then compose for $s$ steps with an RDP accountant, until we perform the next attack and update our $\hat{\sigma}_r$. \fedit{To provide a worst-case measure we choose a threshold that maximises $\hat\eps$ under our attack.} %
\fedit{We note that $n$ %
determines bounds on $\hat\eps_r$ and its CIs. With $n=100$, $\hat\eps_r$ is at most $3.89$. %
Similarly, when computing $\hat \eps_r$, we set $\delta = 1/n$.}
\edit{For further accounting details and the algorithm to compute $\hat\eps$ see Appendix \ref{appendix:empirical_priv}}.

\paragraph{Relationship to DP.} \label{sec:rel2dp}

Our threat model and canary design procedure have been chosen to ensure that the adversary is as strong as possible but still constrained by realistic 
assumptions. %
At least three factors of the design process restrict the adversary in practice:

\begin{enumerate}
    \item The adversary only has partial knowledge of the model updates in the form of an aggregated noisy sum $\tilde{u}$. Furthermore, when designing a canary, the adversary uses a design pool to mock clients. In our design, we have been conservative by assuming the adversary has access to heldout data that matches the federated distribution \edit{by constructing the design pool from the test set of our datasets}. \edit{In practice, adversaries can form design pools from public datasets that match the task of the target model. See Appendix \ref{appendix:limitations} for further discussion.}
    \item The optimization process induces two possible sources of error: the convergence itself and the generalization ability of the canaries to unseen model updates.
    \item We calculate the maximum accuracy that can be derived from the attack \fedit{and use the threshold that maximises $\hat\eps_r$}. This is conservative, as in practice an adversary would have to calibrate a threshold $\gamma$ on a public heldout dataset. Furthermore, they would not be able to perform $n$ fake rounds with the same canary.
\end{enumerate}

If these practical constraints were not imposed on an adversary, then the privacy measure we derive from our attack would essentially be tight with the DP upper bound. For example, if we could always design a perfect canary $\can$ that has gradient orthogonal to all possible model updates $u_i$ then the DP noise $\sigma$ completely determines whether we can distinguish if $\can$ is present in $\tilde{u}$ or not. However, in practice, the assumptions listed above make the attack more difficult as we will see in Section \ref{sec:exp:cifar10}.

\section{Experiments}\label{sec:exp}

\begin{figure*}[t]
   \subfloat[$\sigma=0$\label{fig:cifar10:hist1} ]{%
      \includegraphics[width=0.32\textwidth]{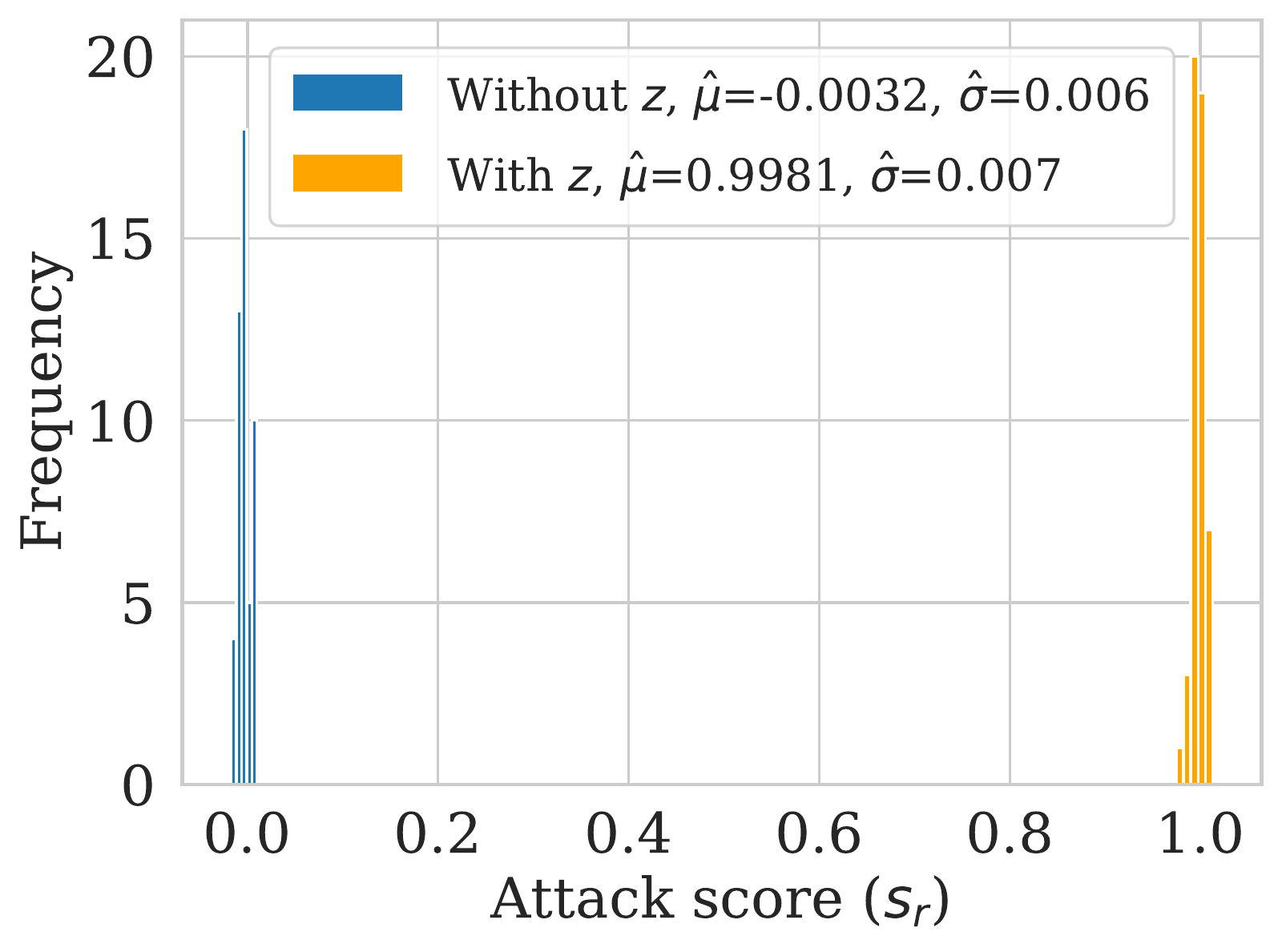}}
\hspace{\fill}
   \subfloat[$\sigma=0.423$\label{fig:cifar10:hist2}]{%
      \includegraphics[width=0.315\textwidth]{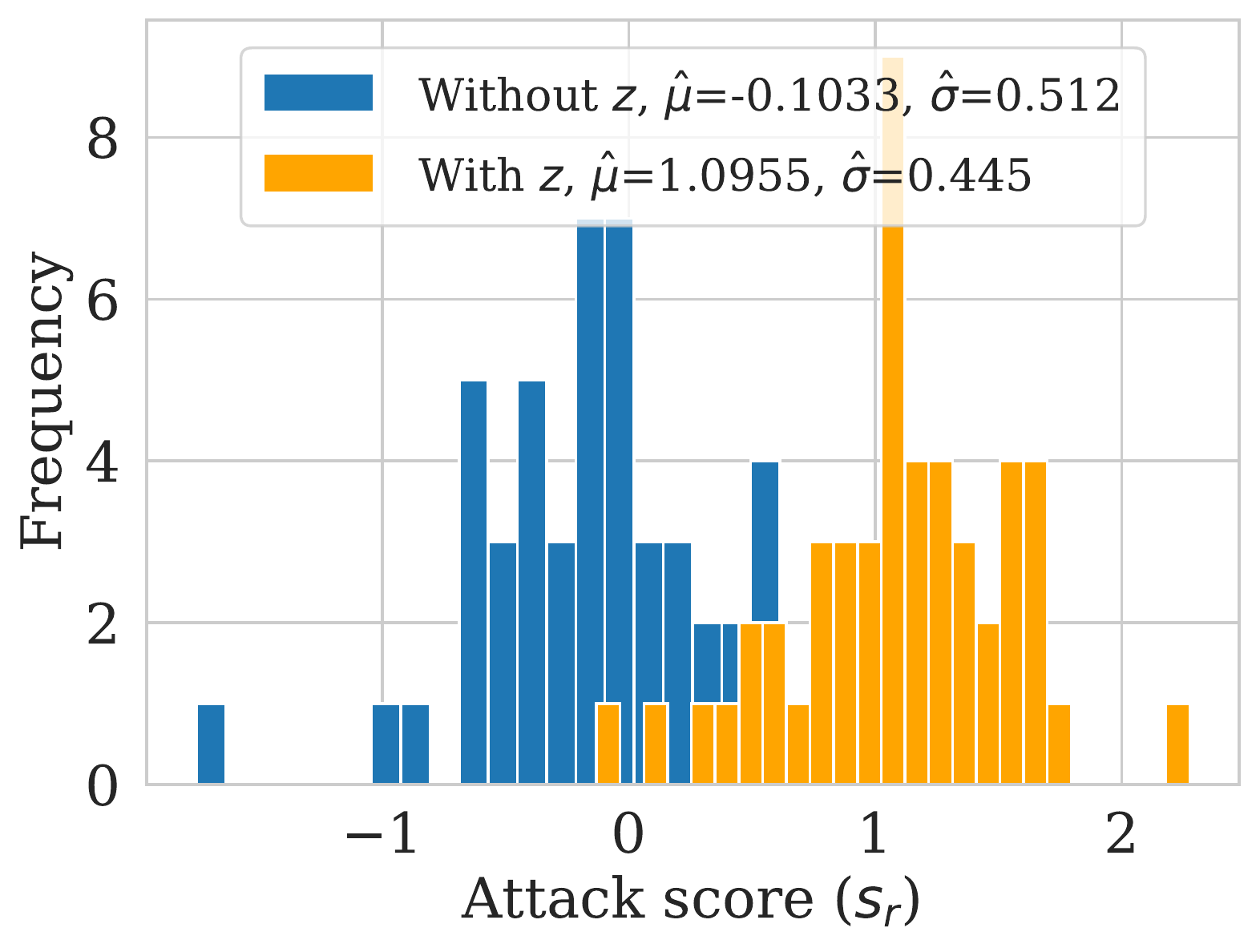}}
\hspace{\fill}
   \subfloat[\label{fig:cifar10:pool}]{%
      \includegraphics[width=0.345\textwidth]{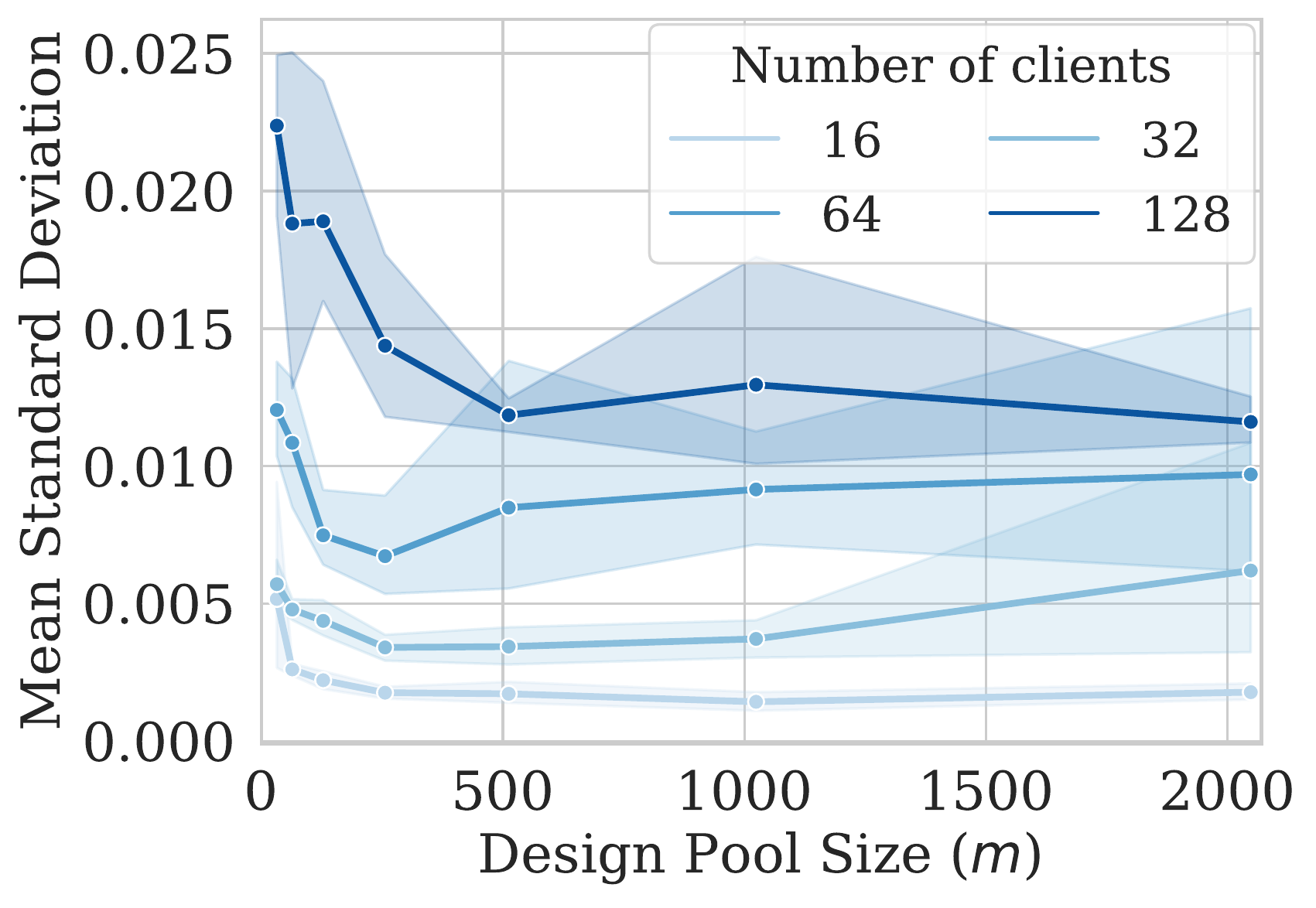}}
    \\
\caption{Experiments on CIFAR-10 on a toy setup where each client holds a single data sample. (a) Attack histograms without DP noise, showing that our attack completely succeeds. (b) Attack histograms with DP noise: as the DP noise $\sigma$ increases, it becomes harder to separate both histograms. (c) Varying the design pool size and the number of participating clients. The $y$-axis shows the mean standard deviation of the histograms formed from the attack.}
\label{fig:cifar10}
\end{figure*}

In this section, we begin by highlighting the power of our attack on the simplest federated setting where participants only have a single sample (Section~\ref{sec:exp:cifar10}). We then show how our attack allows us to monitor privacy empirically over a federated training run (Section~\ref{sec:exp:monitor}) before concluding with an ablation study that shows our canary is robust to various design choices (Section~\ref{sec:exp:ablation}). We open-source the code for \method design and test to reproduce our results~\footnote{Code available at \url{https://github.com/facebookresearch/canife}}.

\paragraph{Setup.}
We evaluate our attack on both image and language models. We utilise LEAF \citep{caldas2018leaf} which provides benchmark federated datasets for simulating clients with non-IID data and a varying number of local samples.
We study image classification on CIFAR10 (IID) \citep{krizhevsky2009learning} and CelebA (non-IID) \citep{liu2015faceattributes}. We train a simple Convolutional Neural Network (CNN) and a ResNet18 model. For our language tasks, we train an LSTM model on non-IID splits of Sent140 %
~\citep{go2009twitter} and Shakespeare %
~\citep{mcmahan2017communication}. For more information on datasets, model architectures and training hyperparameters, see Appendix \ref{appendix:arch}. We train, design, and evaluate our attacks using the FLSim framework\footnote{\url{https://github.com/facebookresearch/FLSim}}. All experiments were run on a single A100 40GB GPU with model training taking at most a few hours. We discuss CPU benchmarks for canary design in Section \ref{sec:exp:ablation}. For canary design, we use the Adam optimizer \citep{kingma2014adam} with learning rate $\beta = 1$ and fix $C=1$. We form the \method design pool from a LEAF test split, resulting in canaries designed on \edit{non-IID mock clients which approximates the training distribution}. \edit{We have clients perform a single local epoch in all experiments, but see Appendix \ref{appendix:limitations} for possible extensions to multiple local epochs. From now, we refer to \say{epoch} as one pass of the federated dataset (in expectation).} \edit{For privacy accounting, we utilise the RDP accountant with subsampling \citep{mironov2019r} implemented via the Opacus library \citep{yousefpour2021opacus}, sampling clients uniformly at each round (i.e., Poisson sampling), see Appendix \ref{appendix:empirical_priv} for more details. }

\subsection{Example Attack: CIFAR10}\label{sec:exp:cifar10}
We first investigate the simplest federated setting where each client holds a single sample. It follows that the clipped model update $u_i$ is simply a scaled gradient of the client's single sample. This corresponds to the central setting where the number of clients per round is equivalent to a (central) batch size. In Figure \ref{fig:cifar10}, we train a ResNet18 model on CIFAR10 to $60\%$ test accuracy and perform our attack. We design a single canary and have the server perform $n=100$ mock rounds where the model is frozen, with half having the canary client inserted and half without. We compute attack scores from these $n=100$ trials and form histograms.

In Figure \ref{fig:cifar10:hist1}, we present one histogram of the attack scores on a model that has 64 clients participating per round with no DP ($\eps = \infty$). We use a design pool of $m=512$ to design the canary. We observe that our attack is quite tight in this non-private setting. Since $\sigma = 0$, we hope that if the canary was designed well, the standard deviation of the histograms would also be close to 0. It turns out that the average standard deviation of the histograms is $0.006$. Recall in Section \ref{sec:rel2dp}, we discussed there is inherent error from both the optimization procedure and the fact the server designs the canary on heldout data but in this case the error is small. In Figure \ref{fig:cifar10:hist2}, we display another example attack, this time for a model trained under privacy with a final $\eps = 25$ corresponding to $\sigma = 0.423$ with 64 clients per round. 
We observe the attack is still fairly tight as the standard deviation of the histograms (average $0.478$) is close to that of the privacy noise $\sigma$.

Finally, we explore how both the design pool size and the number of clients affect the standard deviation of the histograms (Figure \ref{fig:cifar10:pool}). We vary both the number of clients and the design pool size.
We train a model without privacy for each combination of parameters until it reaches 60\% train accuracy and then attack the model, plotting the average standard deviation of the histograms.
We conclude with two observations. First, the attack is much tighter when there is a smaller number of clients participating per round. Second, the size of the design pool has a diminishing effect on reducing the standard deviation. We further explore these in Section \ref{sec:exp:ablation}.

\subsection{Monitoring Empirical Privacy}\label{sec:exp:monitor}
\begin{figure*}[t]
   \subfloat[\fedit{Sent140 (2-layer LSTM)} \label{fig:monitor:sent140} ]{%
      \includegraphics[width=0.31\textwidth]{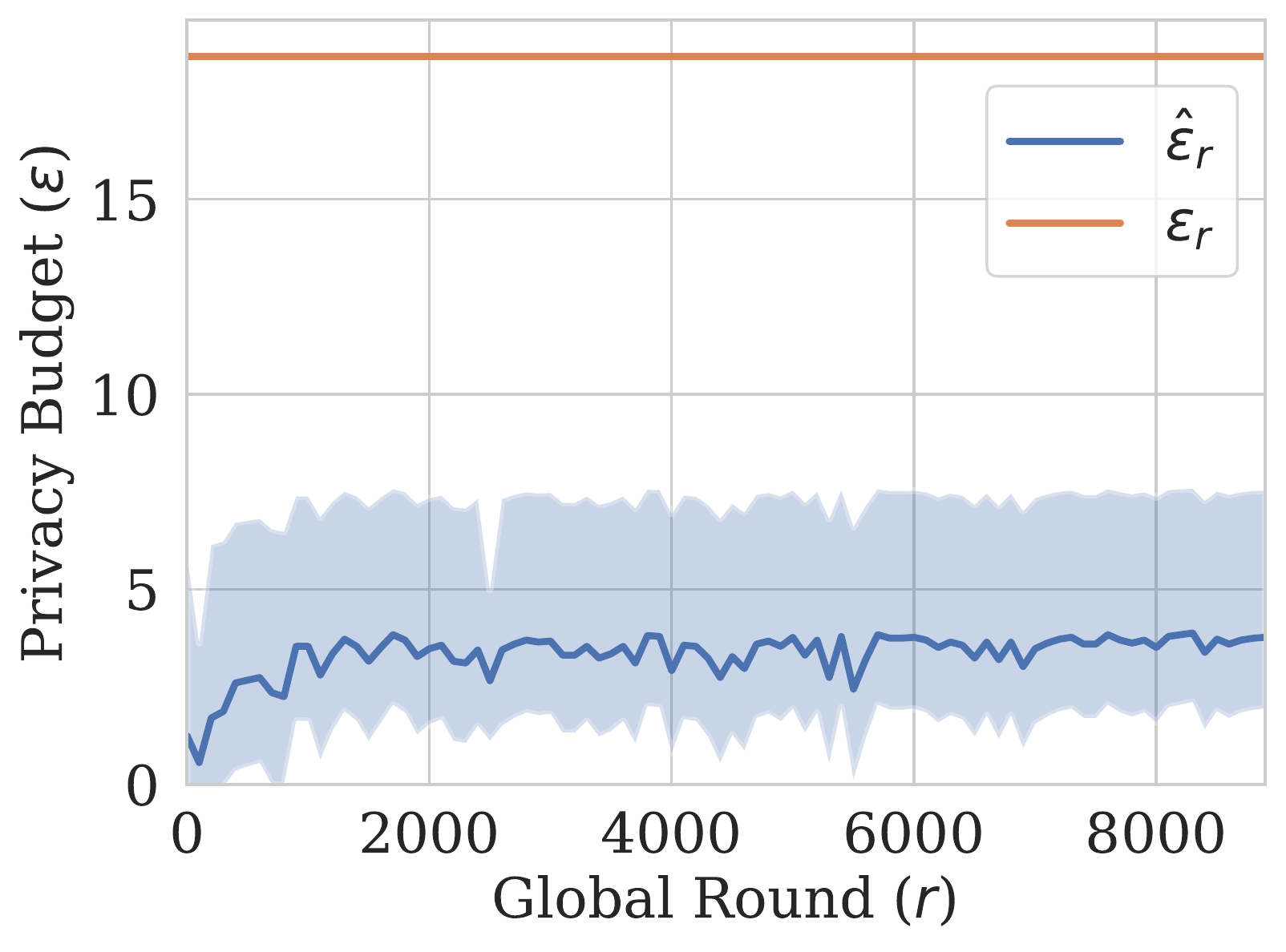}}
\hspace{\fill}
    \subfloat[\fedit{CelebA (ResNet18)} \label{fig:monitor:celeba}]{%
      \includegraphics[width=0.33\textwidth]{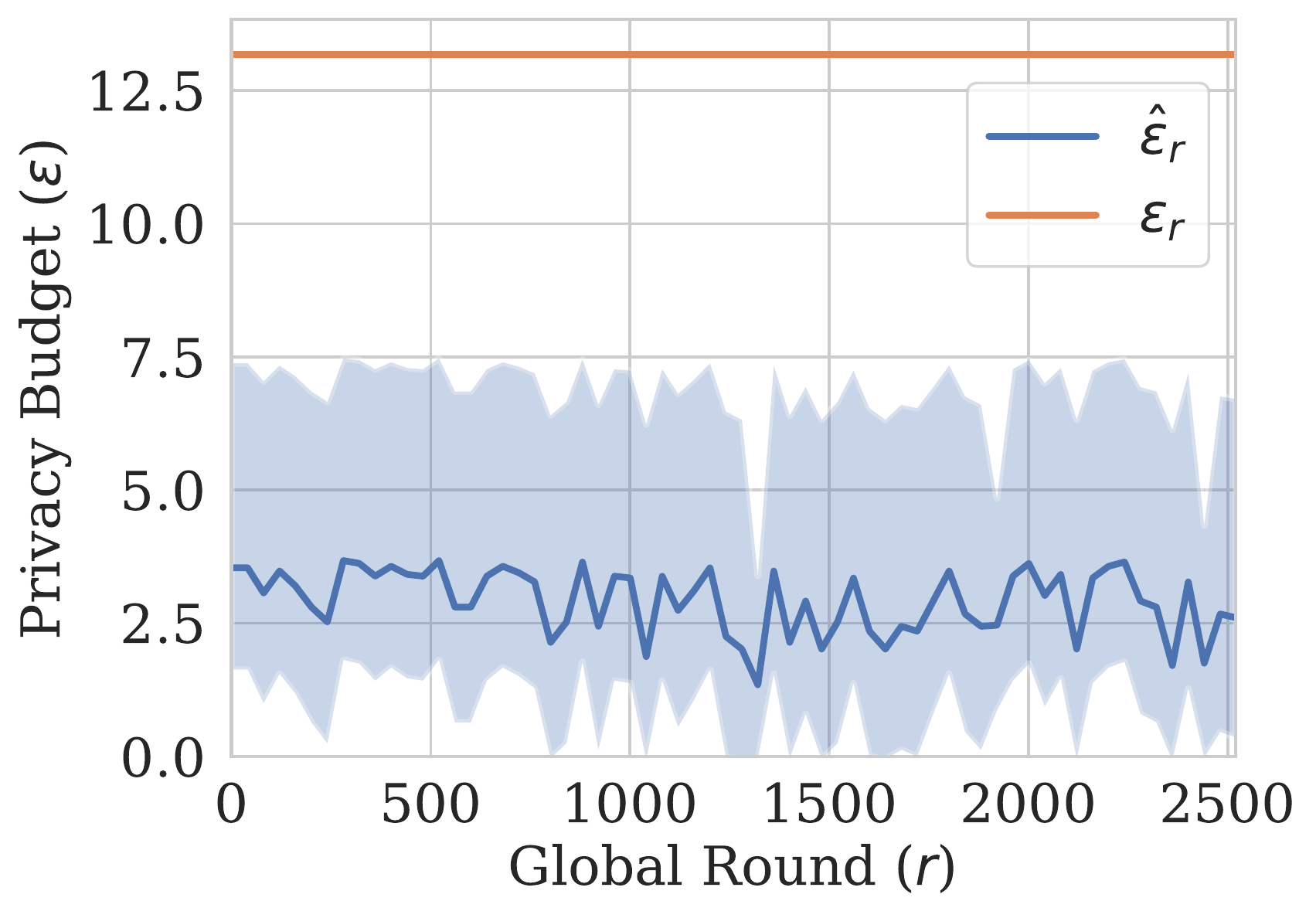}}
\hspace{\fill}
   \subfloat[\fedit{\method empirical epsilon $\hat\eps$}\label{fig:monitor:tasks} ]{%
      \includegraphics[width=0.315\textwidth]{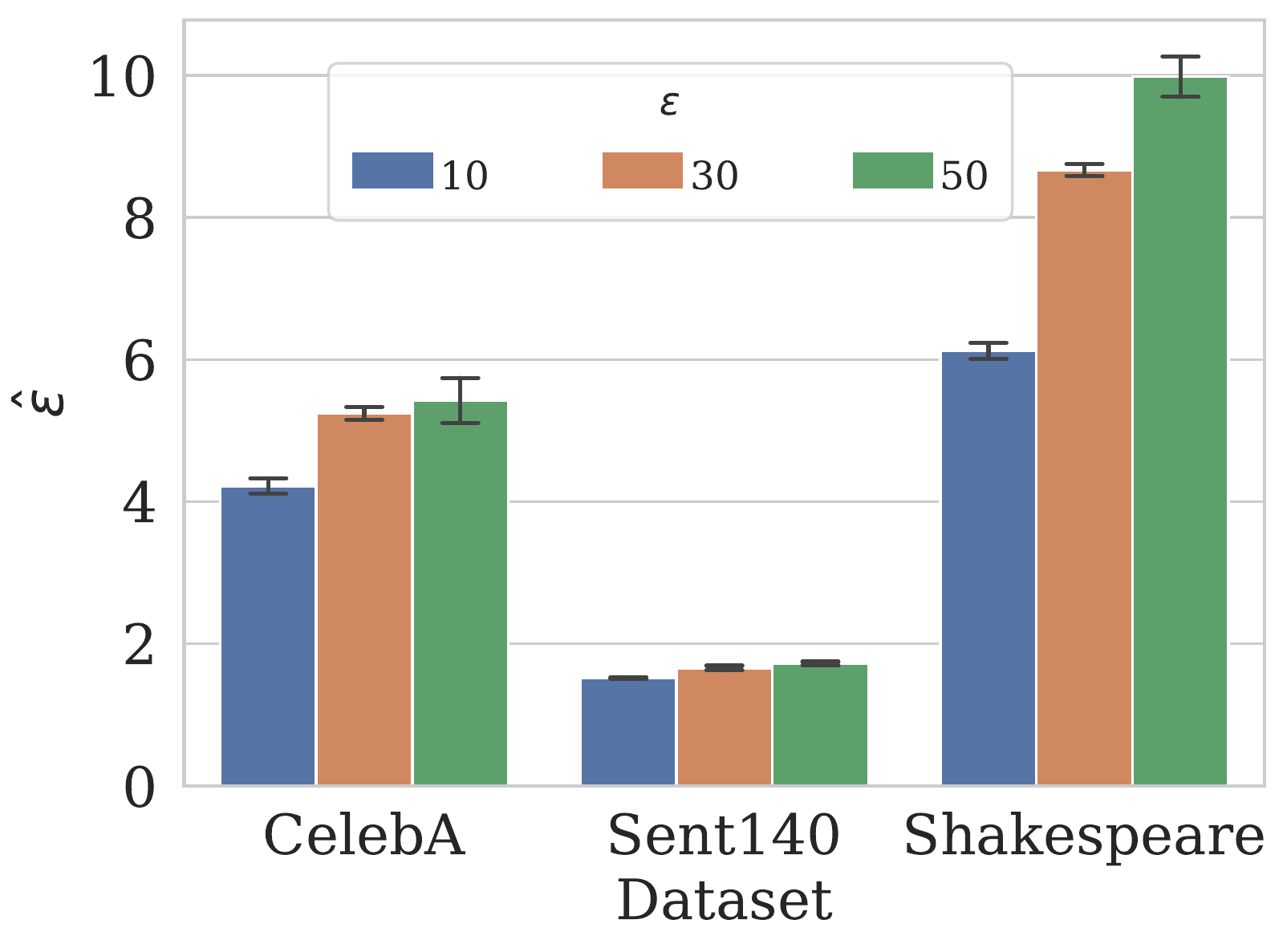}}
      \\
\caption{(a) and (b) Monitoring per-round empirical privacy on LEAF benchmarks. Models are trained to $\eps = 50$ with $100$ clients per round. %
CIs for $\hat{\eps}_r$ are compared with the theoretical per-round epsilon $\eps_r$, rounds with an upper CI of $\infty$ do not have CIs displayed. (c) \method empirical privacy incurred during the whole training, averaged over 5 independent runs (see Figure~\ref{fig:train_and_freeze} for details).}
\label{fig:monitor}
\end{figure*}

\looseness=-1
One main goal of our attack is to provide a lightweight method for monitoring empirical privacy during federated training. We explore this in Figure \ref{fig:monitor}, training ResNet18 on CelebA and a 2-layer LSTM model on Sent140 and Shakespeare. %
We train to a final $\varepsilon = 50$ and achieve $63.9\%$ test accuracy on Sent140, $89.9\%$ on CelebA and $44.8\%$ on Shakespeare. We carry out the \method attack at a regular interval, %
freezing the model, designing a single canary, and performing $n=100$ attack trials with the designed canary before continuing training. This generates scores which are used to compute an empirical per-round privacy measure $\hat{\eps}_r$ %
as outlined in Section \ref{sec:privacy_measure} (see also Appendix~\ref{appendix:monitor_details} for further details). 
The canary sample $\can$ is initialised randomly and we explore how this affects optimization in Section \ref{sec:exp:ablation}. We note in practice that participants (and the server) would not want to waste rounds with a frozen model\edit{, see Appendix \ref{appendix:limitations} for possible extensions}. 

We monitor the optimization of the canary loss by computing the \emph{canary health} which measures the percentage improvement of the final canary loss compared to the initial loss. A perfect health of $1$ means the canary achieved minimal design loss. 
As an example, for Sent140, we find that the average canary health across 5 separate training runs is $0.970 \pm 0.098$. Excluding the first 1000 rounds from each run, the average canary health becomes $0.990 \pm 0.01$. Thus, after two model epochs, optimization stabilises and the initial canary loss is reduced by $99\%$.

In Figure \ref{fig:monitor:sent140}, we display per-round privacy estimates $\hat{\eps}_r$ and their 95\% confidence intervals (CIs) for Sent140 trained with $\eps = 50$ and compare to the (constant) theoretical per-round privacy $\varepsilon_r$. We observe $\hat{\eps}_r$ is initially small and grows to an almost constant level within a single epoch and stays there during training. \fedit{This results in a $4.5\times$ gap between the theoretical $\eps_r$ and $\hat\eps_r$ measured by our attack}. We obtain similar \fedit{$4$ -- $5\times$} gaps for CelebA (ResNet18) in Figure \ref{fig:monitor:celeba} and Shakespeare in Appendix \ref{appendix:empirical_priv:shakes}. We compound these per-round estimates $\hat{\eps}_r$ to provide a cumulative %
measure of privacy $\hat{\eps}$ and display this in Figure \ref{fig:train_and_freeze} for models trained with $\eps \in \{10, 30, 50\}$. Again, there is a significant gap between the final theoretical privacy and the measure derived under \method. The final %
$\hat\eps$ averaged over $5$ separate training runs for each tasks are shown in Figure \ref{fig:monitor:tasks}. \fedit{We note significant gaps between tasks, most notably with Sent140. This is likely due to the small sample rate which provides significant privacy amplification.} We also observe \method estimates are stable across runs with relatively small standard deviation. %

\subsection{Ablation Study}\label{sec:exp:ablation}

\begin{figure*}[t]   \subfloat[Canary Initialisation \label{fig:ablation:init} ]{%
      \includegraphics[width=0.32\textwidth]{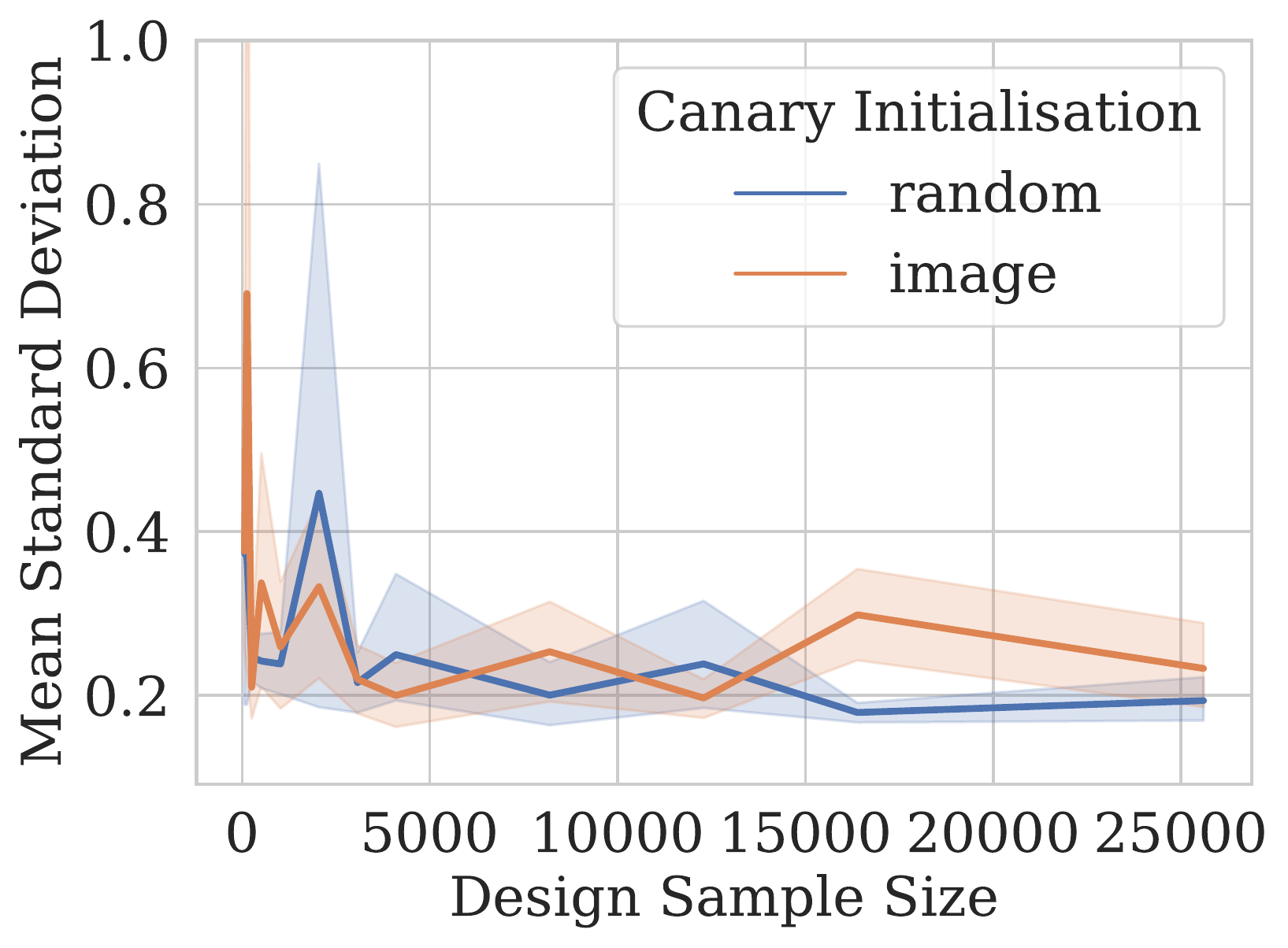}}
\hspace{\fill}
   \subfloat[Design Iterations: Varying $T$ \label{fig:ablation:iters}]{%
      \includegraphics[width=0.32\textwidth]{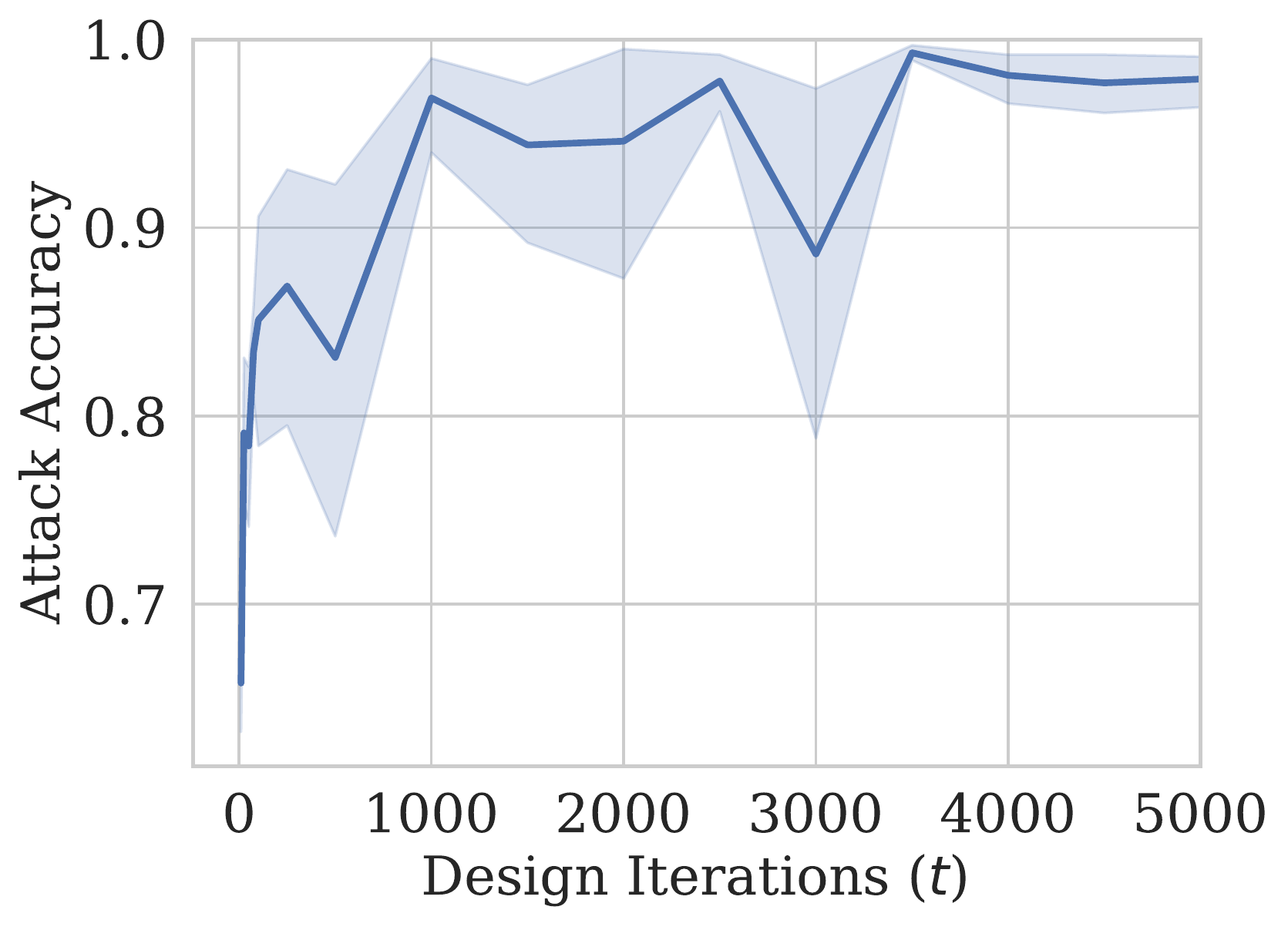}}
\hspace{\fill}
    \subfloat[Model test vs attack accuracy \label{fig:ablation:acc}]{%
      \includegraphics[width=0.315\textwidth]{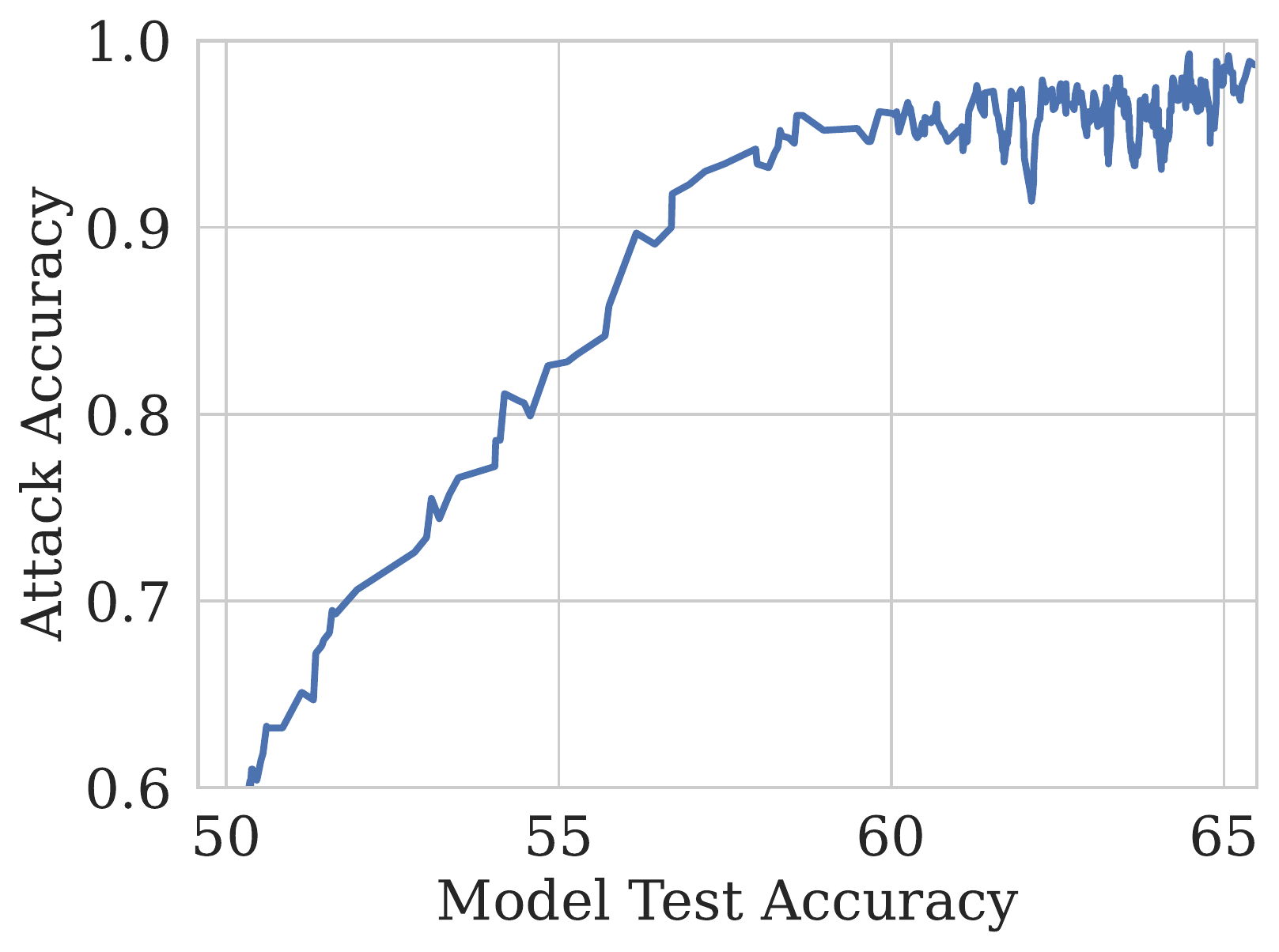}}
      \\
\caption{Ablations studies for \method \textbf{without addition of DP noise}. (a) Canary initialization when varying the design pool size (CelebA, 4-layer CNN). (b) Number of design iterations required to produce robust and well-designed canaries (CelebA, 4-layer CNN). (c) Model vs attack accuracy across 5 runs (Sent140, 2-layer LSTM). Attack accuracy increases as model test accuracy increases.}
\label{fig:ablation}
\end{figure*}

\paragraph{Canary Initialisation.} In Figure \ref{fig:ablation:init}, we explore how initialising the canary affects the optimization. We consider two initialisation strategies: initialising the canary randomly or initialising the canary as a sample from the design pool (which is then excluded from the design). We observe no significant difference on the average standard deviation of the attack histograms.

\paragraph{Design Pool Size.}  In Figure \ref{fig:ablation:init}, we vary the design pool size. We observe there is no significant effect on the average standard deviation of the attack histograms. This confirms what we observed in Figure \ref{fig:cifar10:pool}: the design pool size has diminishing impact on reducing the standard deviation.

\paragraph{Design Iterations.} In Figure \ref{fig:ablation:iters}, we explore how the number of design iterations impacts the quality of the canary measured through the calibrated attack accuracy.  We observe that  just $t=1000$ iterations are needed to obtain $95\%$ accuracy and  that with $t=3000$ the attack improves to almost $100\%$ accuracy, staying close to constant as $t$ increases further. We additionally benchmarked the average CPU time on an M1 MacBook Air (2020) for a single design iteration. For Sent140, it takes an average of 0.06s per iteration and 0.23s for CelebA (ResNet18). For $t=2500$, the design procedure took on average 165s and 591s respectively. Hence, the canary design process takes at most 10 minutes on a CPU and only a few minutes with a GPU. Thus our attack is lightweight, requiring only a few thousand design iterations to achieve near-optimal canaries.

\paragraph{Model Accuracy.} We conclude by noting the accuracy of the model and that of our attack are highly correlated. In Figure \ref{fig:ablation:acc}, we plot both model test accuracy and calibrated attack accuracy across Sent140 trained without DP. We observe early in training, when model accuracy is low, that the attack accuracy is similarly low. Once the model converges to a sufficiently high test accuracy the attack accuracy is close to $100\%$. In experiments that require comparison across different models (e.g., Figure \ref{fig:cifar10:pool}) we checkpoint to a fixed test accuracy to avoid this confounding effect.

\section{Conclusion}

Motivated by the fact that DP is conservative, we consider a more realistic threat model to extract private information about client data when training a model with FL. 
We introduce \method, a novel method to measure empirical privacy where a rogue client crafts a canary sample that results in an outlier model update. We argue that the difficulty of tracing this malicious model update in the aggregated noisy model update provides a tighter measurement of the model's privacy exposure. We hope this work can benefit practical deployments of DP-FL pipelines %
by complementing the theoretical bound with an arguably more realistic measure of the privacy leakage.

\bibliography{iclr2023_conference}

\begin{thebibliography}{58}
\providecommand{\natexlab}[1]{#1}
\providecommand{\url}[1]{\texttt{#1}}
\expandafter\ifx\csname urlstyle\endcsname\relax
  \providecommand{\doi}[1]{doi: #1}\else
  \providecommand{\doi}{doi: \begingroup \urlstyle{rm}\Url}\fi

\bibitem[Abadi et~al.(2016)Abadi, Chu, Goodfellow, McMahan, Mironov, Talwar,
  and Zhang]{abadi2016deep}
Martin Abadi, Andy Chu, Ian Goodfellow, H~Brendan McMahan, Ilya Mironov, Kunal
  Talwar, and Li~Zhang.
\newblock Deep learning with differential privacy.
\newblock In \emph{Proceedings of the 2016 ACM SIGSAC conference on computer
  and communications security}, pp.\  308--318, 2016.

\bibitem[Alistarh et~al.(2017)Alistarh, Grubic, Li, Tomioka, and
  Vojnovic]{alistarh2017qsgd}
Dan Alistarh, Demjan Grubic, Jerry Li, Ryota Tomioka, and Milan Vojnovic.
\newblock {QSGD}: Communication-efficient {SGD} via gradient quantization and
  encoding.
\newblock \emph{Advances in neural information processing systems}, 30, 2017.

\bibitem[Balle et~al.(2020)Balle, Barthe, Gaboardi, Hsu, and
  Sato]{balle2020hypothesis}
Borja Balle, Gilles Barthe, Marco Gaboardi, Justin Hsu, and Tetsuya Sato.
\newblock Hypothesis testing interpretations and renyi differential privacy.
\newblock In \emph{International Conference on Artificial Intelligence and
  Statistics}, pp.\  2496--2506. PMLR, 2020.

\bibitem[Balle et~al.(2022)Balle, Cherubin, and Hayes]{balle2022reconstructing}
Borja Balle, Giovanni Cherubin, and Jamie Hayes.
\newblock Reconstructing training data with informed adversaries.
\newblock \emph{arXiv preprint arXiv:2201.04845}, 2022.

\bibitem[Bassily et~al.(2014)Bassily, Smith, and
  Thakurta]{bassily2014differentially}
Raef Bassily, Adam Smith, and Abhradeep Thakurta.
\newblock Differentially private empirical risk minimization: Efficient
  algorithms and tight error bounds, 2014.

\bibitem[Bell et~al.(2020)Bell, Bonawitz, Gasc{\'o}n, Lepoint, and
  Raykova]{bell2020secure}
James~Henry Bell, Kallista~A Bonawitz, Adri{\`a} Gasc{\'o}n, Tancr{\`e}de
  Lepoint, and Mariana Raykova.
\newblock Secure single-server aggregation with (poly) logarithmic overhead.
\newblock In \emph{Proceedings of the 2020 ACM SIGSAC Conference on Computer
  and Communications Security}, pp.\  1253--1269, 2020.

\bibitem[Biggio et~al.(2013)Biggio, Corona, Maiorca, Nelson, {\v{S}}rndi{\'c},
  Laskov, Giacinto, and Roli]{biggio2013evasion}
Battista Biggio, Igino Corona, Davide Maiorca, Blaine Nelson, Nedim
  {\v{S}}rndi{\'c}, Pavel Laskov, Giorgio Giacinto, and Fabio Roli.
\newblock Evasion attacks against machine learning at test time.
\newblock In \emph{Joint European conference on machine learning and knowledge
  discovery in databases}, pp.\  387--402. Springer, 2013.

\bibitem[Boenisch et~al.(2021)Boenisch, Dziedzic, Schuster, Shamsabadi,
  Shumailov, and Papernot]{boenisch2021curious}
Franziska Boenisch, Adam Dziedzic, Roei Schuster, Ali~Shahin Shamsabadi, Ilia
  Shumailov, and Nicolas Papernot.
\newblock When the curious abandon honesty: Federated learning is not private.
\newblock \emph{arXiv preprint arXiv:2112.02918}, 2021.

\bibitem[Bonawitz et~al.(2017)Bonawitz, Ivanov, Kreuter, Marcedone, McMahan,
  Patel, Ramage, Segal, and Seth]{bonawitz2017practical}
Keith Bonawitz, Vladimir Ivanov, Ben Kreuter, Antonio Marcedone, H~Brendan
  McMahan, Sarvar Patel, Daniel Ramage, Aaron Segal, and Karn Seth.
\newblock Practical secure aggregation for privacy-preserving machine learning.
\newblock In \emph{proceedings of the 2017 ACM SIGSAC Conference on Computer
  and Communications Security}, pp.\  1175--1191, 2017.

\bibitem[Caldas et~al.(2018)Caldas, Duddu, Wu, Li, Kone{\v{c}}n{\`y}, McMahan,
  Smith, and Talwalkar]{caldas2018leaf}
Sebastian Caldas, Sai Meher~Karthik Duddu, Peter Wu, Tian Li, Jakub
  Kone{\v{c}}n{\`y}, H~Brendan McMahan, Virginia Smith, and Ameet Talwalkar.
\newblock {LEAF}: A benchmark for federated settings.
\newblock \emph{arXiv preprint arXiv:1812.01097}, 2018.

\bibitem[Carlini et~al.(2019)Carlini, Liu, Erlingsson, Kos, and
  Song]{carlini2019secret}
Nicholas Carlini, Chang Liu, {\'U}lfar Erlingsson, Jernej Kos, and Dawn Song.
\newblock The secret sharer: Evaluating and testing unintended memorization in
  neural networks.
\newblock In \emph{28th USENIX Security Symposium (USENIX Security 19)}, pp.\
  267--284, 2019.

\bibitem[Carlini et~al.(2021)Carlini, Tramer, Wallace, Jagielski, Herbert-Voss,
  Lee, Roberts, Brown, Song, Erlingsson, et~al.]{carlini2021extracting}
Nicholas Carlini, Florian Tramer, Eric Wallace, Matthew Jagielski, Ariel
  Herbert-Voss, Katherine Lee, Adam Roberts, Tom Brown, Dawn Song, Ulfar
  Erlingsson, et~al.
\newblock Extracting training data from large language models.
\newblock In \emph{30th USENIX Security Symposium (USENIX Security 21)}, pp.\
  2633--2650, 2021.

\bibitem[Dwork \& Roth(2014)Dwork and Roth]{dwork2014foundations}
Cynthia Dwork and Aaron Roth.
\newblock The algorithmic foundations of differential privacy.
\newblock \emph{Foundations and Trends in Theoretical Computer Science}, 2014.

\bibitem[Dwork et~al.(2006)Dwork, McSherry, Nissim, and
  Smith]{dwork2006calibrating}
Cynthia Dwork, Frank McSherry, Kobbi Nissim, and Adam Smith.
\newblock Calibrating noise to sensitivity in private data analysis.
\newblock In \emph{Theory of cryptography conference}, pp.\  265--284.
  Springer, 2006.

\bibitem[Fowl et~al.(2021)Fowl, Geiping, Czaja, Goldblum, and
  Goldstein]{fowl2021robbing}
Liam Fowl, Jonas Geiping, Wojtek Czaja, Micah Goldblum, and Tom Goldstein.
\newblock Robbing the fed: Directly obtaining private data in federated
  learning with modified models.
\newblock \emph{arXiv preprint arXiv:2110.13057}, 2021.

\bibitem[Fowl et~al.(2022)Fowl, Geiping, Reich, Wen, Czaja, Goldblum, and
  Goldstein]{fowl2022decepticons}
Liam Fowl, Jonas Geiping, Steven Reich, Yuxin Wen, Wojtek Czaja, Micah
  Goldblum, and Tom Goldstein.
\newblock Decepticons: Corrupted transformers breach privacy in federated
  learning for language models.
\newblock \emph{arXiv preprint arXiv:2201.12675}, 2022.

\bibitem[Frey(2021)]{frey2021introducing}
Suzanne Frey.
\newblock Introducing {Android}’s private compute services.
\newblock 2021.
\newblock URL \url{https://security.googleblog.com/2021/09/}.

\bibitem[Geiping et~al.(2020)Geiping, Bauermeister, Dr{\"o}ge, and
  Moeller]{geiping2020inverting}
Jonas Geiping, Hartmut Bauermeister, Hannah Dr{\"o}ge, and Michael Moeller.
\newblock Inverting gradients---how easy is it to break privacy in federated
  learning?
\newblock \emph{Advances in Neural Information Processing Systems},
  33:\penalty0 16937--16947, 2020.

\bibitem[Go et~al.(2009)Go, Bhayani, and Huang]{go2009twitter}
Alec Go, Richa Bhayani, and Lei Huang.
\newblock Twitter sentiment classification using distant supervision.
\newblock \emph{CS224N project report, Stanford}, 1\penalty0 (12):\penalty0
  2009, 2009.

\bibitem[Gopi et~al.(2021)Gopi, Lee, and Wutschitz]{gopi2021numerical}
Sivakanth Gopi, Yin~Tat Lee, and Lukas Wutschitz.
\newblock Numerical composition of differential privacy.
\newblock \emph{Advances in Neural Information Processing Systems},
  34:\penalty0 11631--11642, 2021.

\bibitem[Guo et~al.(2021)Guo, Sablayrolles, Jégou, and
  Kiela]{guo2021gradientbased}
Chuan Guo, Alexandre Sablayrolles, Hervé Jégou, and Douwe Kiela.
\newblock Gradient-based adversarial attacks against text transformers.
\newblock \emph{arXiv preprint arXiv:2104.13733}, 2021.

\bibitem[Gupta et~al.(2022)Gupta, Huang, Zhong, Gao, Li, and
  Chen]{gupta2022recovering}
Samyak Gupta, Yangsibo Huang, Zexuan Zhong, Tianyu Gao, Kai Li, and Danqi Chen.
\newblock Recovering private text in federated learning of language models,
  2022.

\bibitem[He et~al.(2016)He, Zhang, Ren, and Sun]{he2016deep}
Kaiming He, Xiangyu Zhang, Shaoqing Ren, and Jian Sun.
\newblock Deep residual learning for image recognition.
\newblock In \emph{Proceedings of the IEEE conference on computer vision and
  pattern recognition}, pp.\  770--778, 2016.

\bibitem[Huba et~al.(2022)Huba, Nguyen, Malik, Zhu, Rabbat, Yousefpour, Wu,
  Zhan, Ustinov, Srinivas, et~al.]{huba2022papaya}
Dzmitry Huba, John Nguyen, Kshitiz Malik, Ruiyu Zhu, Mike Rabbat, Ashkan
  Yousefpour, Carole-Jean Wu, Hongyuan Zhan, Pavel Ustinov, Harish Srinivas,
  et~al.
\newblock Papaya: Practical, private, and scalable federated learning.
\newblock \emph{Proceedings of Machine Learning and Systems}, 4:\penalty0
  814--832, 2022.

\bibitem[Jagielski et~al.(2020)Jagielski, Ullman, and
  Oprea]{jagielski2020auditing}
Matthew Jagielski, Jonathan Ullman, and Alina Oprea.
\newblock Auditing differentially private machine learning: {H}ow private is
  private {SGD}?
\newblock \emph{Advances in Neural Information Processing Systems},
  33:\penalty0 22205--22216, 2020.

\bibitem[Jang et~al.(2016)Jang, Gu, and Poole]{jang2016categorical}
Eric Jang, Shixiang Gu, and Ben Poole.
\newblock Categorical reparameterization with {Gumbel-Softmax}.
\newblock \emph{arXiv preprint arXiv:1611.01144}, 2016.

\bibitem[Jayaraman \& Evans(2019)Jayaraman and Evans]{jayaraman2019evaluating}
Bargav Jayaraman and David Evans.
\newblock Evaluating differentially private machine learning in practice.
\newblock In \emph{28th USENIX Security Symposium (USENIX Security 19)}, pp.\
  1895--1912, 2019.

\bibitem[Jeon et~al.(2021)Jeon, Lee, Oh, Ok, et~al.]{jeon2021gradient}
Jinwoo Jeon, Kangwook Lee, Sewoong Oh, Jungseul Ok, et~al.
\newblock Gradient inversion with generative image prior.
\newblock \emph{Advances in Neural Information Processing Systems},
  34:\penalty0 29898--29908, 2021.

\bibitem[Kairouz et~al.(2015)Kairouz, Oh, and
  Viswanath]{kairouz2015composition}
Peter Kairouz, Sewoong Oh, and Pramod Viswanath.
\newblock The composition theorem for differential privacy.
\newblock In \emph{International conference on machine learning}, pp.\
  1376--1385. PMLR, 2015.

\bibitem[Kairouz et~al.(2019)Kairouz, McMahan, Avent, Bellet, Bennis, Bhagoji,
  Bonawitz, Charles, Cormode, Cummings, D'Oliveira, Eichner, Rouayheb, Evans,
  Gardner, Garrett, Gascón, Ghazi, Gibbons, Gruteser, Harchaoui, He, He, Huo,
  Hutchinson, Hsu, Jaggi, Javidi, Joshi, Khodak, Konečný, Korolova,
  Koushanfar, Koyejo, Lepoint, Liu, Mittal, Mohri, Nock, Özgür, Pagh,
  Raykova, Qi, Ramage, Raskar, Song, Song, Stich, Sun, Suresh, Tramèr,
  Vepakomma, Wang, Xiong, Xu, Yang, Yu, Yu, and Zhao]{kairouz2019advances}
Peter Kairouz, H.~Brendan McMahan, Brendan Avent, Aurélien Bellet, Mehdi
  Bennis, Arjun~Nitin Bhagoji, Kallista Bonawitz, Zachary Charles, Graham
  Cormode, Rachel Cummings, Rafael G.~L. D'Oliveira, Hubert Eichner, Salim~El
  Rouayheb, David Evans, Josh Gardner, Zachary Garrett, Adrià Gascón, Badih
  Ghazi, Phillip~B. Gibbons, Marco Gruteser, Zaid Harchaoui, Chaoyang He, Lie
  He, Zhouyuan Huo, Ben Hutchinson, Justin Hsu, Martin Jaggi, Tara Javidi,
  Gauri Joshi, Mikhail Khodak, Jakub Konečný, Aleksandra Korolova, Farinaz
  Koushanfar, Sanmi Koyejo, Tancrède Lepoint, Yang Liu, Prateek Mittal,
  Mehryar Mohri, Richard Nock, Ayfer Özgür, Rasmus Pagh, Mariana Raykova,
  Hang Qi, Daniel Ramage, Ramesh Raskar, Dawn Song, Weikang Song, Sebastian~U.
  Stich, Ziteng Sun, Ananda~Theertha Suresh, Florian Tramèr, Praneeth
  Vepakomma, Jianyu Wang, Li~Xiong, Zheng Xu, Qiang Yang, Felix~X. Yu, Han Yu,
  and Sen Zhao.
\newblock Advances and open problems in federated learning, 2019.

\bibitem[Kairouz et~al.(2021)Kairouz, McMahan, Song, Thakkar, Thakurta, and
  Xu]{kairouz2021practical}
Peter Kairouz, Brendan McMahan, Shuang Song, Om~Thakkar, Abhradeep Thakurta,
  and Zheng Xu.
\newblock Practical and private (deep) learning without sampling or shuffling.
\newblock In \emph{International Conference on Machine Learning}, pp.\
  5213--5225. PMLR, 2021.

\bibitem[Karimireddy et~al.(2020{\natexlab{a}})Karimireddy, Jaggi, Kale, Mohri,
  Reddi, Stich, and Suresh]{karimireddy2020mime}
Sai~Praneeth Karimireddy, Martin Jaggi, Satyen Kale, Mehryar Mohri, Sashank~J
  Reddi, Sebastian~U Stich, and Ananda~Theertha Suresh.
\newblock Mime: Mimicking centralized stochastic algorithms in federated
  learning.
\newblock \emph{arXiv preprint arXiv:2008.03606}, 2020{\natexlab{a}}.

\bibitem[Karimireddy et~al.(2020{\natexlab{b}})Karimireddy, Kale, Mohri, Reddi,
  Stich, and Suresh]{karimireddy2020scaffold}
Sai~Praneeth Karimireddy, Satyen Kale, Mehryar Mohri, Sashank Reddi, Sebastian
  Stich, and Ananda~Theertha Suresh.
\newblock Scaffold: Stochastic controlled averaging for federated learning.
\newblock In \emph{International Conference on Machine Learning}, pp.\
  5132--5143. PMLR, 2020{\natexlab{b}}.

\bibitem[Kingma \& Ba(2014)Kingma and Ba]{kingma2014adam}
Diederik~P Kingma and Jimmy Ba.
\newblock Adam: A method for stochastic optimization.
\newblock \emph{arXiv preprint arXiv:1412.6980}, 2014.

\bibitem[Krizhevsky et~al.(2009)Krizhevsky, Hinton,
  et~al.]{krizhevsky2009learning}
Alex Krizhevsky, Geoffrey Hinton, et~al.
\newblock Learning multiple layers of features from tiny images.
\newblock 2009.

\bibitem[Liu et~al.(2015)Liu, Luo, Wang, and Tang]{liu2015faceattributes}
Ziwei Liu, Ping Luo, Xiaogang Wang, and Xiaoou Tang.
\newblock Deep learning face attributes in the wild.
\newblock In \emph{Proceedings of International Conference on Computer Vision
  (ICCV)}, December 2015.

\bibitem[Mahloujifar et~al.(2022)Mahloujifar, Sablayrolles, Cormode, and
  Jha]{mahloujifar2022optimal}
Saeed Mahloujifar, Alexandre Sablayrolles, Graham Cormode, and Somesh Jha.
\newblock Optimal membership inference bounds for adaptive composition of
  sampled gaussian mechanisms, 2022.

\bibitem[McMahan et~al.(2017{\natexlab{a}})McMahan, Moore, Ramage, Hampson, and
  y~Arcas]{mcmahan2017communication}
Brendan McMahan, Eider Moore, Daniel Ramage, Seth Hampson, and Blaise~Aguera
  y~Arcas.
\newblock Communication-efficient learning of deep networks from decentralized
  data.
\newblock In \emph{Artificial intelligence and statistics}, pp.\  1273--1282.
  PMLR, 2017{\natexlab{a}}.

\bibitem[McMahan et~al.(2017{\natexlab{b}})McMahan, Ramage, Talwar, and
  Zhang]{mcmahan2017learning}
H~Brendan McMahan, Daniel Ramage, Kunal Talwar, and Li~Zhang.
\newblock Learning differentially private recurrent language models.
\newblock \emph{arXiv preprint arXiv:1710.06963}, 2017{\natexlab{b}}.

\bibitem[Mironov(2017)]{mironov2017renyi}
Ilya Mironov.
\newblock R{\'e}nyi differential privacy.
\newblock In \emph{2017 IEEE 30th computer security foundations symposium
  (CSF)}, pp.\  263--275. IEEE, 2017.

\bibitem[Mironov et~al.(2019)Mironov, Talwar, and Zhang]{mironov2019r}
Ilya Mironov, Kunal Talwar, and Li~Zhang.
\newblock R$\backslash$'enyi differential privacy of the sampled gaussian
  mechanism.
\newblock \emph{arXiv preprint arXiv:1908.10530}, 2019.

\bibitem[Nasr et~al.(2019)Nasr, Shokri, and Houmansadr]{nasr2019comprehensive}
Milad Nasr, Reza Shokri, and Amir Houmansadr.
\newblock Comprehensive privacy analysis of deep learning: Passive and active
  white-box inference attacks against centralized and federated learning.
\newblock In \emph{2019 IEEE symposium on security and privacy (SP)}, pp.\
  739--753. IEEE, 2019.

\bibitem[Nasr et~al.(2021)Nasr, Songi, Thakurta, Papemoti, and
  Carlin]{nasr2021adversary}
Milad Nasr, Shuang Songi, Abhradeep Thakurta, Nicolas Papemoti, and Nicholas
  Carlin.
\newblock Adversary instantiation: Lower bounds for differentially private
  machine learning.
\newblock In \emph{2021 IEEE Symposium on Security and Privacy (SP)}, pp.\
  866--882. IEEE, 2021.

\bibitem[Parikh et~al.(2022)Parikh, Dupuy, and Gupta]{parikh2022canary}
Rahil Parikh, Christophe Dupuy, and Rahul Gupta.
\newblock Canary extraction in natural language understanding models.
\newblock \emph{arXiv preprint arXiv:2203.13920}, 2022.

\bibitem[Paszke et~al.(2019)Paszke, Gross, Massa, Lerer, Bradbury, Chanan,
  Killeen, Lin, Gimelshein, Antiga, Desmaison, Kopf, Yang, DeVito, Raison,
  Tejani, Chilamkurthy, Steiner, Fang, Bai, and Chintala]{paske2019pytorch}
Adam Paszke, Sam Gross, Francisco Massa, Adam Lerer, James Bradbury, Gregory
  Chanan, Trevor Killeen, Zeming Lin, Natalia Gimelshein, Luca Antiga, Alban
  Desmaison, Andreas Kopf, Edward Yang, Zachary DeVito, Martin Raison, Alykhan
  Tejani, Sasank Chilamkurthy, Benoit Steiner, Lu~Fang, Junjie Bai, and Soumith
  Chintala.
\newblock {PyTorch}: An imperative style, high-performance deep learning
  library.
\newblock In \emph{Advances in Neural Information Processing Systems 32}. 2019.

\bibitem[Ramaswamy et~al.(2020)Ramaswamy, Thakkar, Mathews, Andrew, McMahan,
  and Beaufays]{ramaswamy2020training}
Swaroop Ramaswamy, Om~Thakkar, Rajiv Mathews, Galen Andrew, H~Brendan McMahan,
  and Fran{\c{c}}oise Beaufays.
\newblock Training production language models without memorizing user data.
\newblock \emph{arXiv preprint arXiv:2009.10031}, 2020.

\bibitem[Reddi et~al.(2020)Reddi, Charles, Zaheer, Garrett, Rush,
  Kone{\v{c}}n{\`y}, Kumar, and McMahan]{reddi2020adaptive}
Sashank Reddi, Zachary Charles, Manzil Zaheer, Zachary Garrett, Keith Rush,
  Jakub Kone{\v{c}}n{\`y}, Sanjiv Kumar, and H~Brendan McMahan.
\newblock Adaptive federated optimization.
\newblock \emph{arXiv preprint arXiv:2003.00295}, 2020.

\bibitem[Sablayrolles et~al.(2019)Sablayrolles, Douze, Schmid, Ollivier, and
  J{\'e}gou]{sablayrolles2019white}
Alexandre Sablayrolles, Matthijs Douze, Cordelia Schmid, Yann Ollivier, and
  Herv{\'e} J{\'e}gou.
\newblock White-box vs black-box: Bayes optimal strategies for membership
  inference.
\newblock In \emph{International Conference on Machine Learning}, pp.\
  5558--5567. PMLR, 2019.

\bibitem[Shi et~al.(2022)Shi, Chen, Zhang, Jia, and Yu]{shi2022just}
Weiyan Shi, Si~Chen, Chiyuan Zhang, Ruoxi Jia, and Zhou Yu.
\newblock Just fine-tune twice: Selective differential privacy for large
  language models.
\newblock \emph{arXiv preprint arXiv:2204.07667}, 2022.

\bibitem[Shokri et~al.(2017)Shokri, Stronati, Song, and
  Shmatikov]{shokri2017membership}
Reza Shokri, Marco Stronati, Congzheng Song, and Vitaly Shmatikov.
\newblock Membership inference attacks against machine learning models.
\newblock In \emph{2017 IEEE symposium on security and privacy (SP)}, pp.\
  3--18. IEEE, 2017.

\bibitem[Stock et~al.(2022)Stock, Shilov, Mironov, and
  Sablayrolles]{stock2022defending}
Pierre Stock, Igor Shilov, Ilya Mironov, and Alexandre Sablayrolles.
\newblock Defending against reconstruction attacks with {R}{\'e}nyi
  differential privacy.
\newblock \emph{arXiv preprint arXiv:2202.07623}, 2022.

\bibitem[Thakkar et~al.(2020)Thakkar, Ramaswamy, Mathews, and
  Beaufays]{thakkar2020understanding}
Om~Thakkar, Swaroop Ramaswamy, Rajiv Mathews, and Fran{\c{c}}oise Beaufays.
\newblock Understanding unintended memorization in federated learning.
\newblock \emph{arXiv preprint arXiv:2006.07490}, 2020.

\bibitem[Wang et~al.(2019{\natexlab{a}})Wang, Sahu, Yang, Joshi, and
  Kar]{wang2019matcha}
Jianyu Wang, Anit~Kumar Sahu, Zhouyi Yang, Gauri Joshi, and Soummya Kar.
\newblock Matcha: Speeding up decentralized {SGD} via matching decomposition
  sampling.
\newblock In \emph{2019 Sixth Indian Control Conference (ICC)}, pp.\  299--300.
  IEEE, 2019{\natexlab{a}}.

\bibitem[Wang et~al.(2019{\natexlab{b}})Wang, Balle, and
  Kasiviswanathan]{wang2019subsampled}
Yu-Xiang Wang, Borja Balle, and Shiva~Prasad Kasiviswanathan.
\newblock Subsampled {R}{\'e}nyi differential privacy and analytical moments
  accountant.
\newblock In \emph{The 22nd International Conference on Artificial Intelligence
  and Statistics}, pp.\  1226--1235. PMLR, 2019{\natexlab{b}}.

\bibitem[Wen et~al.(2022)Wen, Geiping, Fowl, Goldblum, and
  Goldstein]{wen2022fishing}
Yuxin Wen, Jonas Geiping, Liam Fowl, Micah Goldblum, and Tom Goldstein.
\newblock Fishing for user data in large-batch federated learning via gradient
  magnification.
\newblock \emph{arXiv preprint arXiv:2202.00580}, 2022.

\bibitem[Xu et~al.(2022)Xu, Song, Tian, Agrawal, Granqvist, van Dalen, Zhang,
  Argueta, Han, Deng, et~al.]{xu2022training}
Mingbin Xu, Congzheng Song, Ye~Tian, Neha Agrawal, Filip Granqvist, Rogier van
  Dalen, Xiao Zhang, Arturo Argueta, Shiyi Han, Yaqiao Deng, et~al.
\newblock Training large-vocabulary neural language models by private federated
  learning for resource-constrained devices.
\newblock \emph{arXiv preprint arXiv:2207.08988}, 2022.

\bibitem[Yin et~al.(2021)Yin, Mallya, Vahdat, Alvarez, Kautz, and
  Molchanov]{yin2021see}
Hongxu Yin, Arun Mallya, Arash Vahdat, Jose~M Alvarez, Jan Kautz, and Pavlo
  Molchanov.
\newblock See through gradients: Image batch recovery via gradinversion.
\newblock In \emph{Proceedings of the IEEE/CVF Conference on Computer Vision
  and Pattern Recognition}, pp.\  16337--16346, 2021.

\bibitem[Yousefpour et~al.(2021)Yousefpour, Shilov, Sablayrolles, Testuggine,
  Prasad, Malek, Nguyen, Ghosh, Bharadwaj, Zhao, et~al.]{yousefpour2021opacus}
Ashkan Yousefpour, Igor Shilov, Alexandre Sablayrolles, Davide Testuggine,
  Karthik Prasad, Mani Malek, John Nguyen, Sayan Ghosh, Akash Bharadwaj,
  Jessica Zhao, et~al.
\newblock Opacus: User-friendly differential privacy library in {PyTorch}.
\newblock \emph{arXiv preprint arXiv:2109.12298}, 2021.

\end{thebibliography}
\bibliographystyle{iclr2023_conference}

\appendix
\newpage 
\renewcommand{\canarygrad}{\nabla \ell(\can)}

\section{Connections to likelihood ratio test}
\label{appendix:lkelihood}
Let us assume that each model update follows a Gaussian distribution $\gaussian(\mu, \Sigma)$.
The sum of $k$ model updates then either follows $\gaussian(k \mu, k \Sigma)$ (without the canary) or $\gaussian(k \mu + \canarygrad, k \Sigma)$ (with the canary), recalling that $u_c \propto \canarygrad$. Then we have
\begin{align*}
    p_0(u) &= \frac{1}{\sqrt{\det(2 \pi k \Sigma)}} \exp \left( - (u - k \mu)^T (k \Sigma)^{-1} (u - k \mu) / 2 \right)\\
    p_1(u) &= \frac{1}{\sqrt{\det(2 \pi k \Sigma)}} \exp \left( - (u - (k \mu + \canarygrad))^T (k \Sigma)^{-1} (u - (k \mu + \canarygrad)) / 2 \right) \\
     &= \frac{1}{\sqrt{\det(2 \pi k \Sigma)}} \exp \left( - ((u - k \mu) - \canarygrad))^T (k \Sigma)^{-1} ((u - k \mu) - \canarygrad) / 2 \right).
\end{align*}
We can write the (log) likelihood ratio as
\begin{align*}
    \log \frac{p_1(u)}{p_0(u)} &= \frac{1}{2} \left( (u - k \mu)^T (k \Sigma)^{-1} (u - k \mu) - ((u - k \mu) - \canarygrad))^T (k \Sigma)^{-1} ((u - k \mu) - \canarygrad) \right) \\
    &= \frac{1}{2} \left( (u - k \mu)^T (k \Sigma)^{-1} (u - k \mu) - ((u - k \mu) - \canarygrad))^T (k \Sigma)^{-1} ((u - k \mu) - \canarygrad) \right) \\
    &= \canarygrad^T (k \Sigma)^{-1} (u - k \mu) - \frac{1}{2} \canarygrad^T (k \Sigma)^{-1} \canarygrad.
\end{align*}

In particular for the centers of the Gaussian with and without the canary, $u \in \{ k\mu, k \mu + \canarygrad\}$, 
\begin{align*}
    \log \left( \frac{p_1(u)}{p_0(u)} \right) &= \pm \frac{1}{2} \canarygrad^T (k \Sigma)^{-1} \canarygrad.
\end{align*}
Maximizing this term will thus help separate the two Gaussians. 
However, doing this directly is infeasible as it requires to form and invert the full covariance matrix $\Sigma$ in very high dimensions. 
Instead, we propose to \emph{minimize} $z \mapsto (\canarygrad^T) \Sigma (\canarygrad)$ as it is tractable and can be done with SGD. Note that for sample model updates $\{u_i\}$ we can estimate the (uncentered) covariance matrix as $ \frac{1}{n} \sum_i u_i u_i^T$ and thus

\begin{align*}
    (\canarygrad^T) \Sigma (\canarygrad) \approx \frac{1}{n} \sum_i \canarygrad^T (u_i u_i^T)\canarygrad = \frac{1}{n} \sum_i \langle u_i, \canarygrad \rangle^2.
\end{align*}
Which, ignoring constants, is the first term of $\mathcal{L}(z)$ defined in Equation~\ref{eq:loss2}. One could alternatively minimise $\langle \canarygrad, \hat{\mu} \rangle^2$ with $\hat{\mu} = \frac{1}{n} \sum_i u_i$. We explore the empirical differences in Appendix \ref{appendix:ablation}.

\section{Datasets \& Model Architectures}\label{appendix:arch}

Here we detail the training setup and model architectures for our experiments. In all experiments we train with \textsc{DP-FedSGD} and without momentum. We use both client $(\eta_C)$ and server $(\eta_S)$ learning rates. We train without dropout in all model architectures. In more detail:
\begin{itemize}
    \item \textbf{CIFAR10} is an image classification tasks with $10$ classes \citep{krizhevsky2009learning}. We train a ResNet18 model \citep{he2016deep} on CIFAR10. We form an IID split of $50,000$ train users and $10,000$ test users where each user holds a single sample and thus has a local batch size of $1$. We use a client learning rate of $\eta_C = 0.01$ and server learning rate $\eta_S = 1$.
    \item \textbf{CelebA} is a binary image classification task \citep{liu2015faceattributes}. We train a ResNet18 model with $11,177,538$ parameters and a simple Convolutional Neural Network (CNN) with four convolutional layers following that used by \citet{caldas2018leaf} with $29,282$ parameters. We use the standard non-IID LEAF split resulting in $8408$ train users and $935$ test users and a local batch size of $32$. We train with a client learning rate of $\eta_C = 0.899$ and a server learning rate of $\eta_S = 0.0797$.
    \item \textbf{Sent140} is a sentiment analysis (binary classification) task \citep{go2009twitter}.  We train a 2-layer LSTM with $272,102$ parameters on Sent140 following the architecture of \citet{caldas2018leaf}. We use standard non-IID LEAF splits resulting in $59,214$ train users and $39,477$ test users and a local batch size of $32$. We train for $15$ epochs, and for $\eps=0$ achieve an average test accuracy of $64.8\%$. We train with a client learning rate of $\eta_C = 5.75$ and a server learning rate of $\eta_S = 0.244$.
    \item \textbf{Shakespeare} is a next character prediction task with $47$ classes \citep{mcmahan2017communication}. We train a similar LSTM model to Sent140, based on the architecture used by \citet{caldas2018leaf} with $819,920$ parameters. We use standard non-IID LEAF splits with $1016$ train users and $113$ test users and a local batch size of $128$. We train our models for $15$ epochs resulting in an average final test accuracy of $44.4\%$ for $\eps = 0$. We use a client learning rate of $\eta_C = 3$ and a server learning rate of $\eta_S = 0.524$.
\end{itemize}

\begin{algorithm}[t]
\caption{Measuring $\hat\eps$} \label{alg:measure}
\begin{algorithmic}[1] 
\Input Number of rounds $R$, Privacy parameter $\delta$, Sampling rate $q$, Number of attack scores $n$, Attack frequency $s$
    \For{$r=1, \dots, R$}
        \If{$r\mod s = 0$}
            \State Freeze the model $\theta_r$ and use Algorithm \ref{alg:design} to compute attack scores $\{s_i\}_{i=1}^n$
            \State \fedit{Calculate $\mathrm{FPR}_\gamma$ and $\mathrm{FNR}_\gamma$ from $\{s_i\}$ at each threshold $\gamma \in \{s_1, \dots, s_n\}$}
            \State \fedit{Compute $\hat{\eps}_r \leftarrow \max_{\gamma}\left(\log \frac{1-\delta-\mathrm{FPR}_\gamma}{\mathrm{FNR}_\gamma},\log \frac{1-\delta-\mathrm{FNR}_\gamma}{\mathrm{FPR}_\gamma}\right)$}
            \State \fedit{Compute $\hat{\sigma}_r \leftarrow \text{GetNoise}(\hat\eps_r, \delta)$ \Comment{Estimate one-step noise multiplier}}
            \State \fedit{$\hat\sigma_{r+i} \leftarrow \hat\sigma_r$ for $i \in [1, s)$ \Comment{ $\hat\sigma_r$ is the estimate for rounds $[r, r+s)$}}
        \EndIf
    \EndFor
    \State \Return $\hat{\eps} \leftarrow \text{GetPrivacy}(\{\hat{\sigma}_r\}; \delta, q)$ \Comment{Compose each noise estimate under an RDP accountant with amplification by subsampling using sample rate $q$, \edit{see \citep{mironov2019r}}}
\end{algorithmic}
\end{algorithm}

\begin{figure*}[t]
\centering
  \subfloat[\fedit{Shakespeare per-round privacy ($\eps=50$)}\label{fig:appendix:shakes:eps50}]{%
      \includegraphics[width=0.42\textwidth]{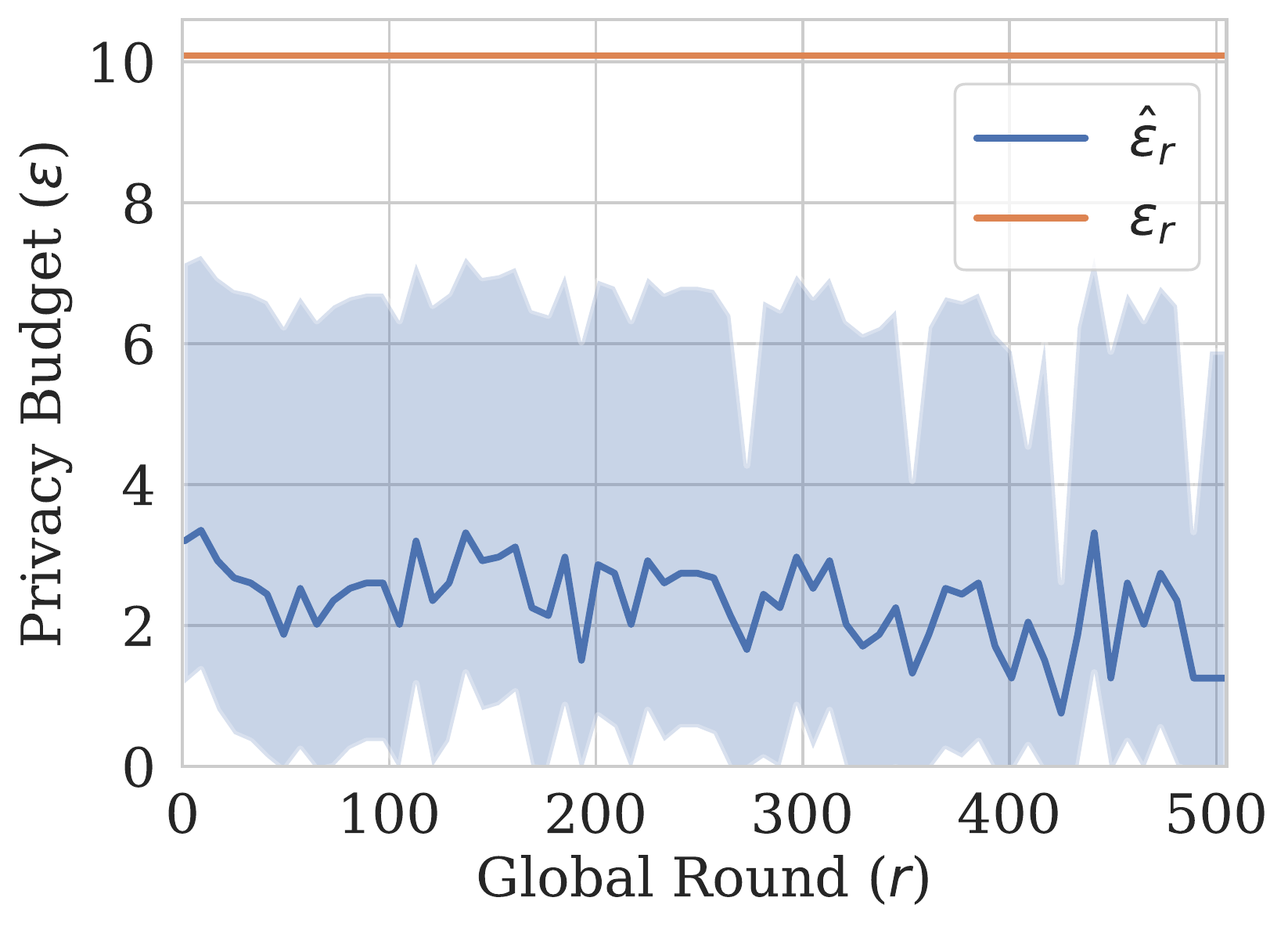}}
  \subfloat[\fedit{Shakespeare per-round privacy ($\eps=\infty$)}\label{fig:appendix:shakes:eps0}]{%
      \includegraphics[width=0.42\textwidth]{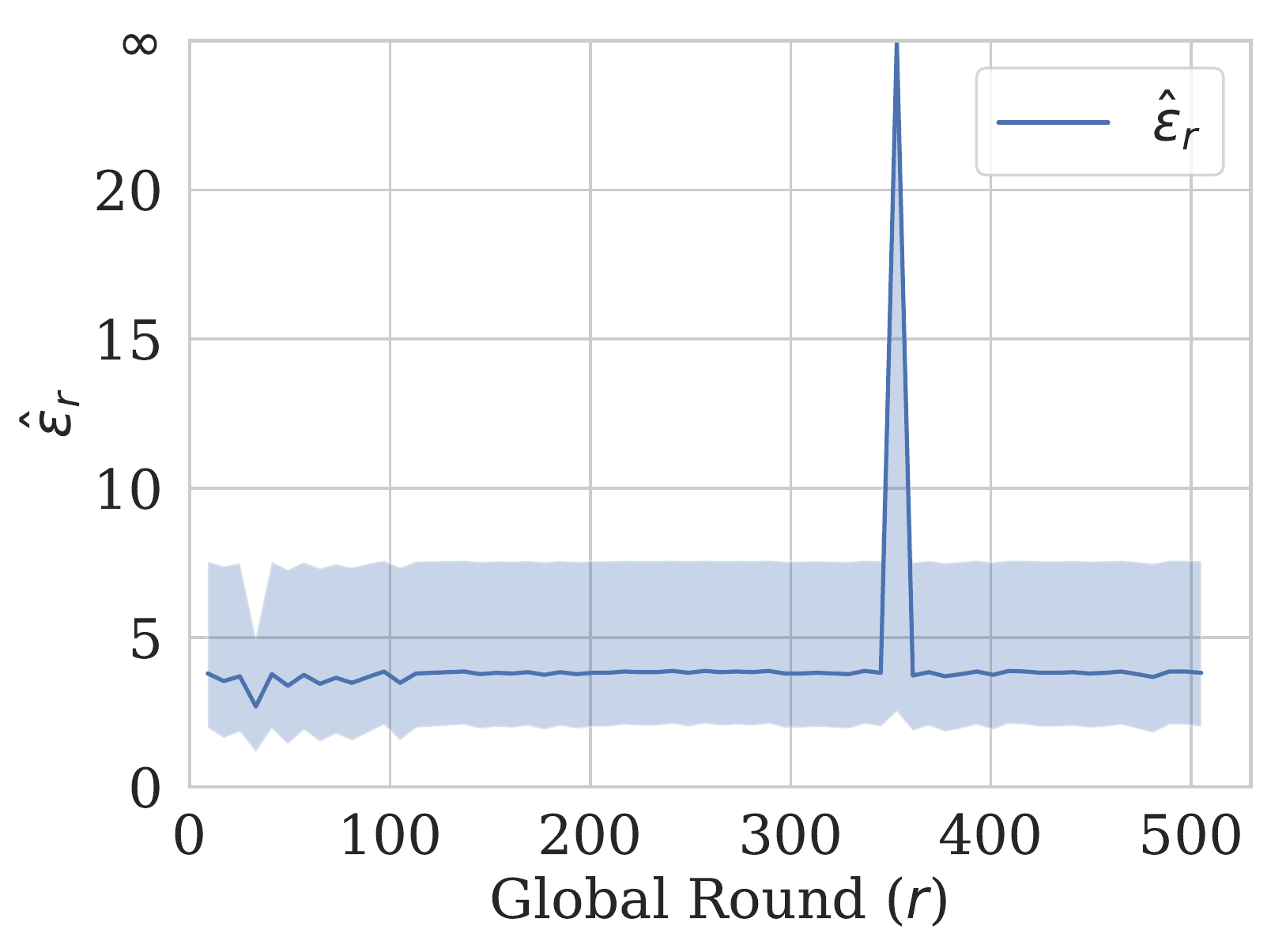}}
    \\
\caption{Measuring privacy during Shakespeare training}
\label{fig:appendix:shakes}
\end{figure*}

\begin{figure*}[t]   
\subfloat[\fedit{Sent140 (2-layer LSTM)}]{%
      \includegraphics[width=0.315\textwidth]{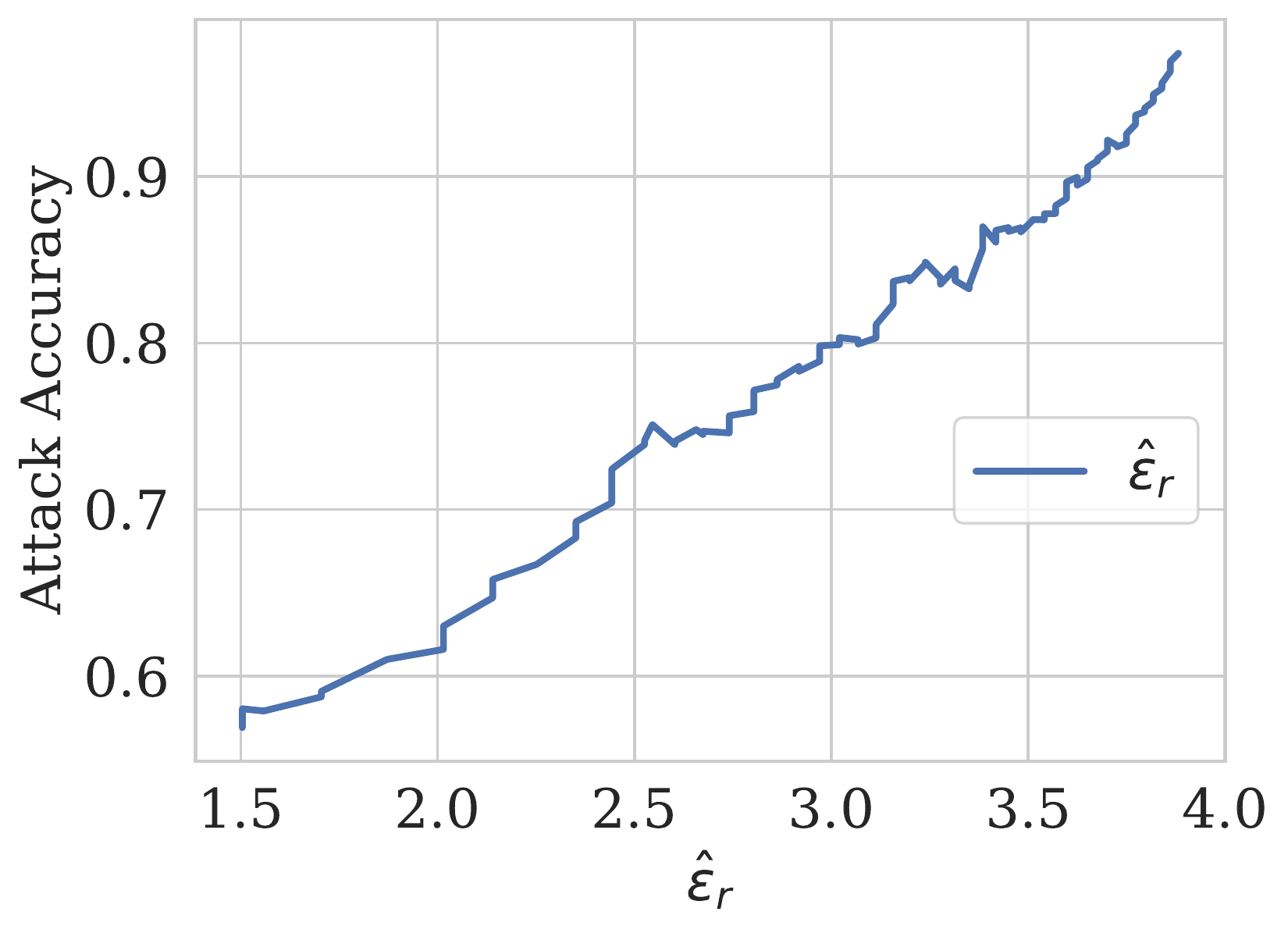}}
\hspace{\fill}
    \subfloat[\fedit{Shakespeare (2-layer LSTM)}]{%
      \includegraphics[width=0.32\textwidth]{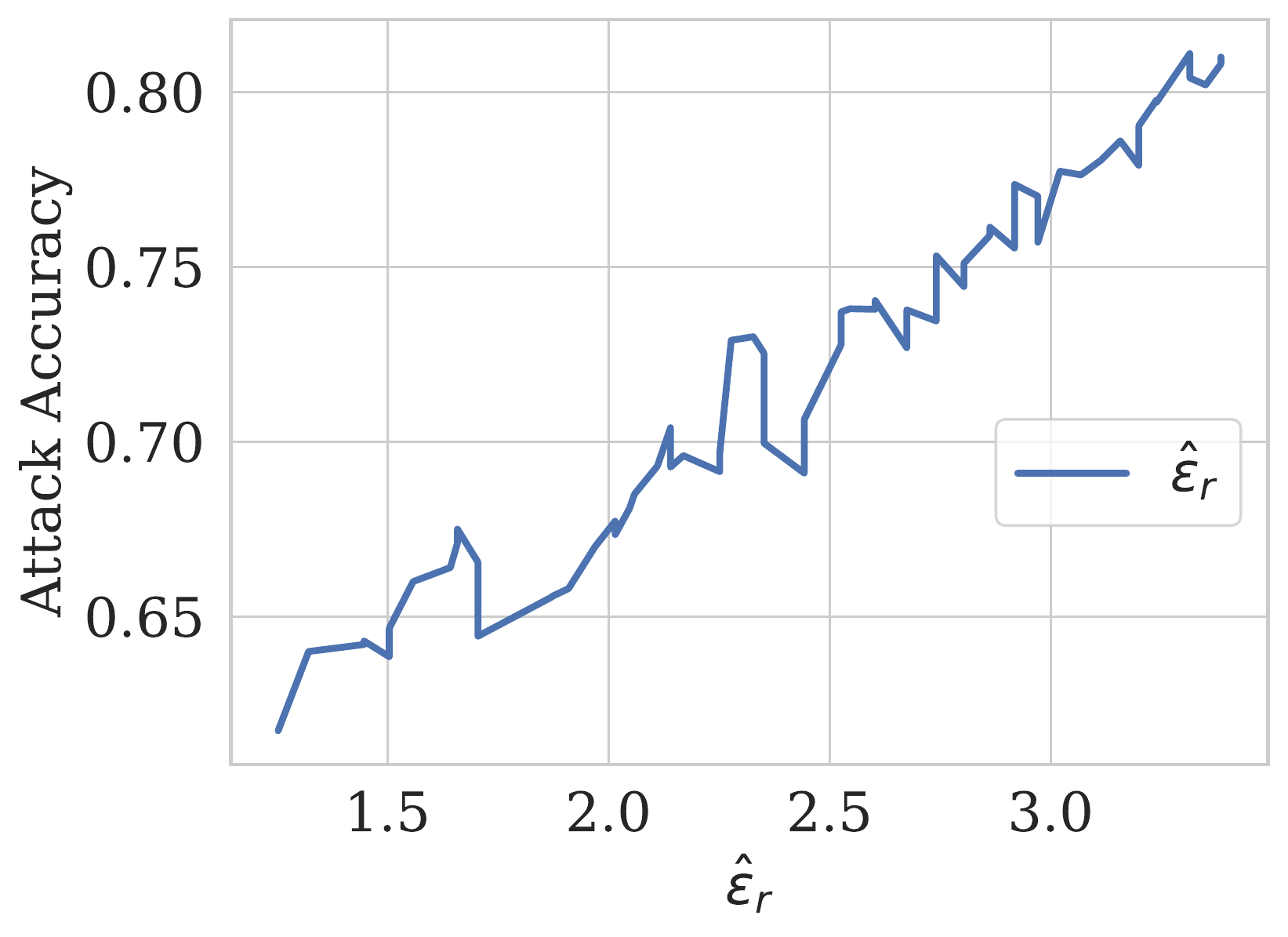}}
\hspace{\fill}
   \subfloat[\fedit{CelebA (ResNet18)}]{%
      \includegraphics[width=0.315\textwidth]{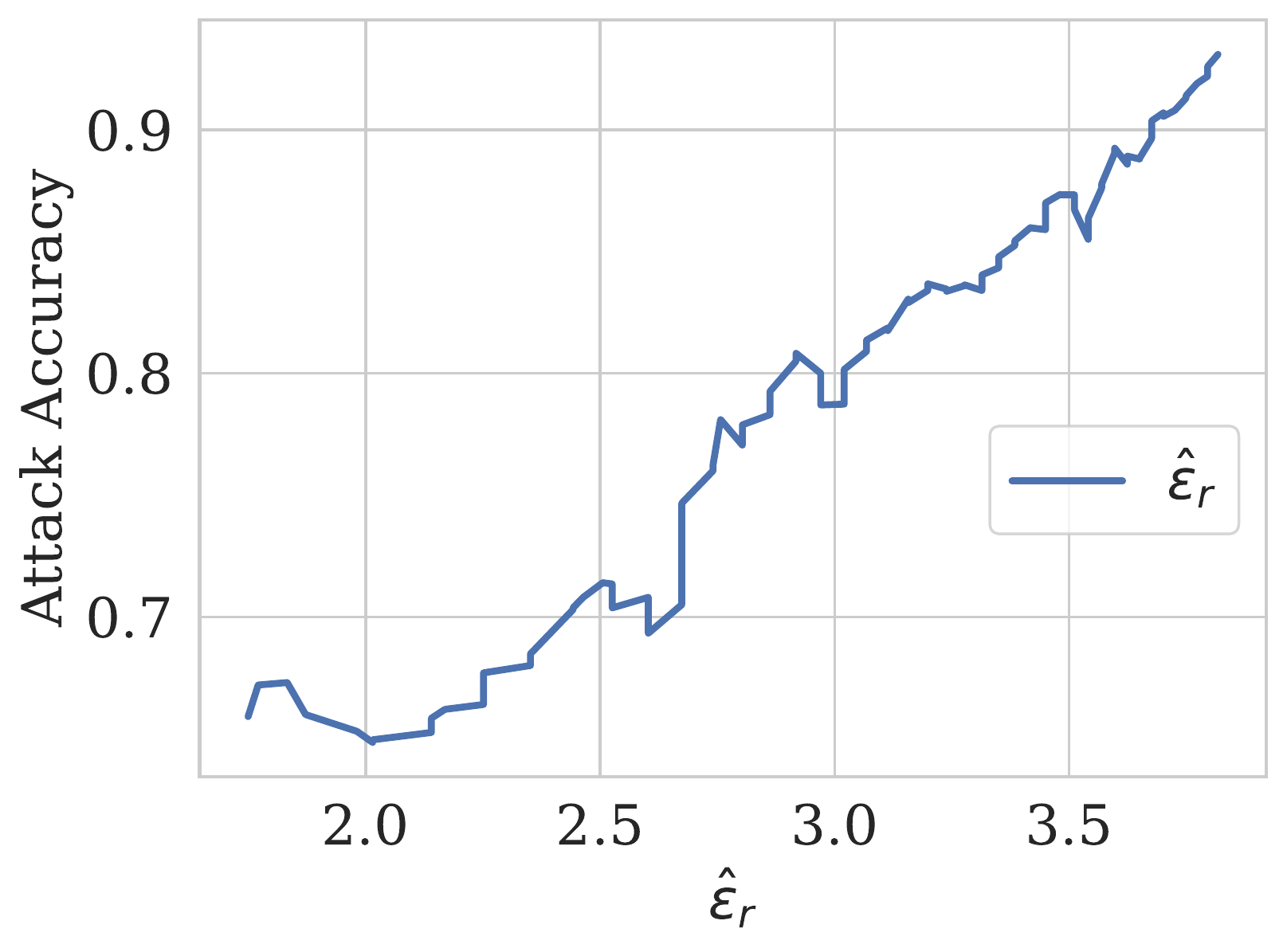}}
      \\
\caption{Per-round $\hat\eps_r$ estimates against attack accuracy. Estimates are computed over 5 training runs with a final theoretical $\eps=50$.}
\label{fig:appendix:acc}
\end{figure*}

\begin{figure*}[t]   \subfloat[Loss modifications \label{fig:appendix:ablation:loss} ]{%
      \includegraphics[width=0.32\textwidth]{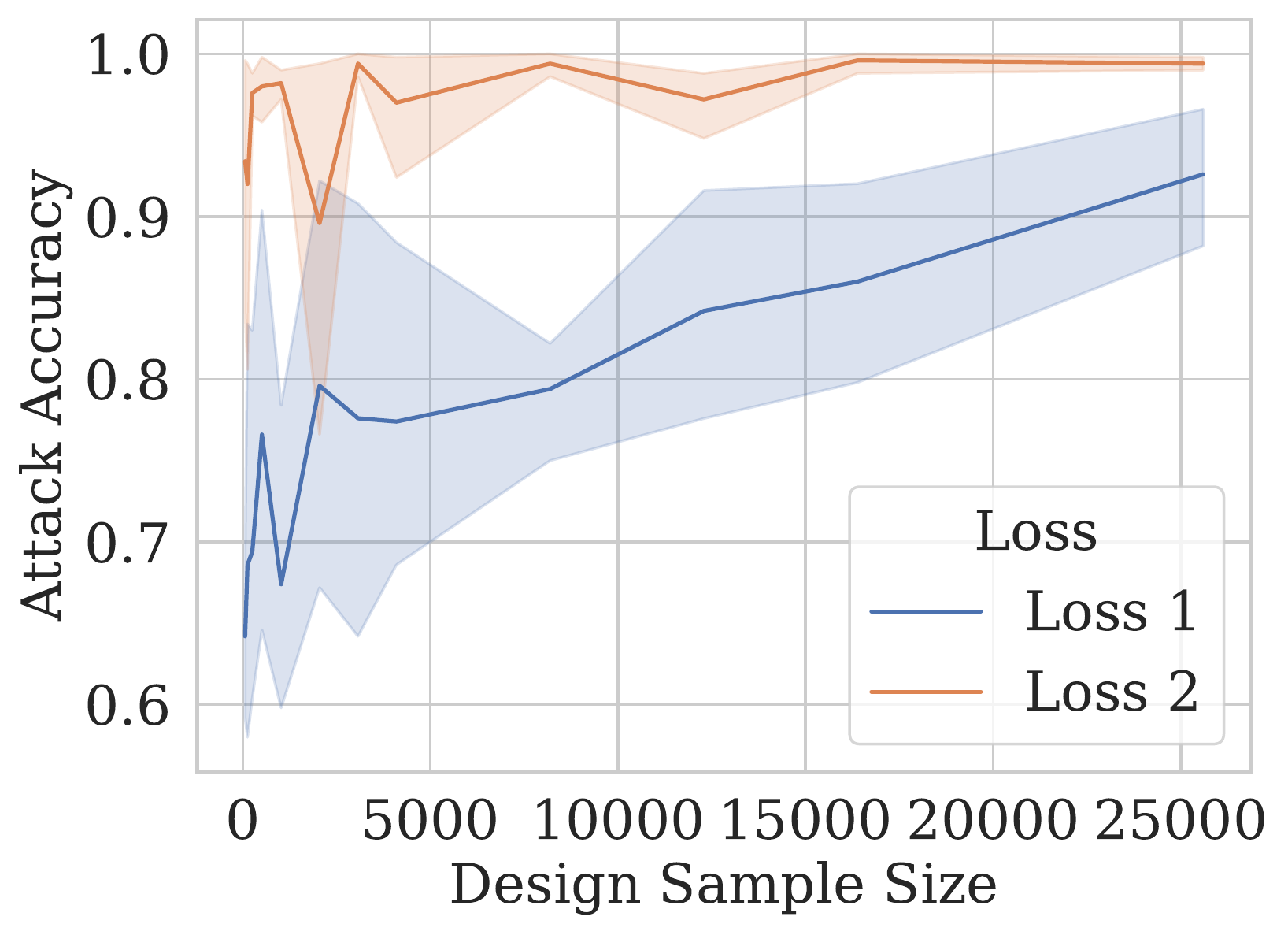}}
\hspace{\fill}
    \subfloat[Varying number of clients per round: Attack accuracy \label{fig:appendix:ablation:client1}]{%
      \includegraphics[width=0.315\textwidth]{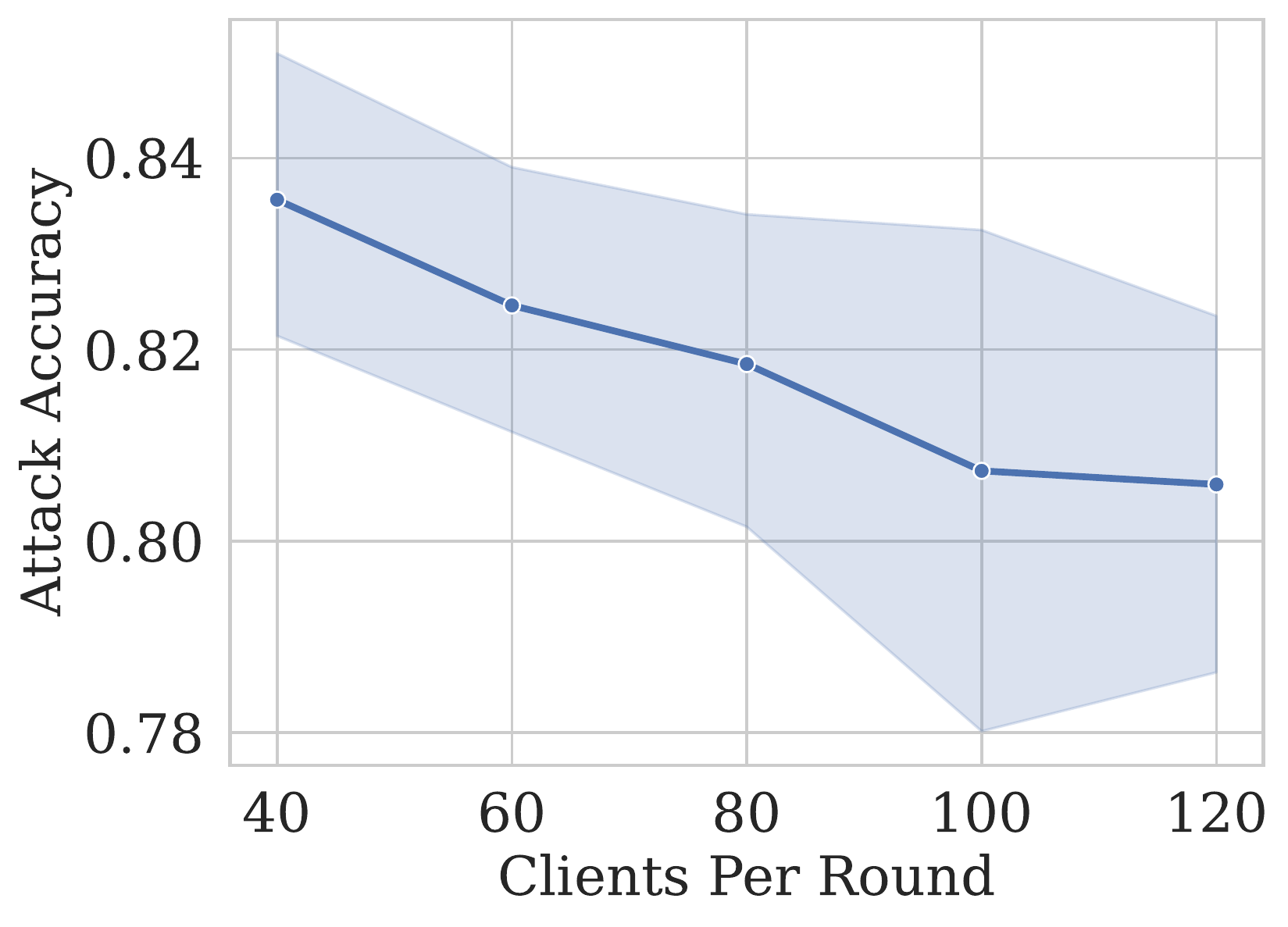}}
\hspace{\fill}
   \subfloat[\fedit{Varying number of clients per round: Empirical $\hat\eps$ \label{fig:appendix:ablation:client2}}]{%
      \includegraphics[width=0.32\textwidth]{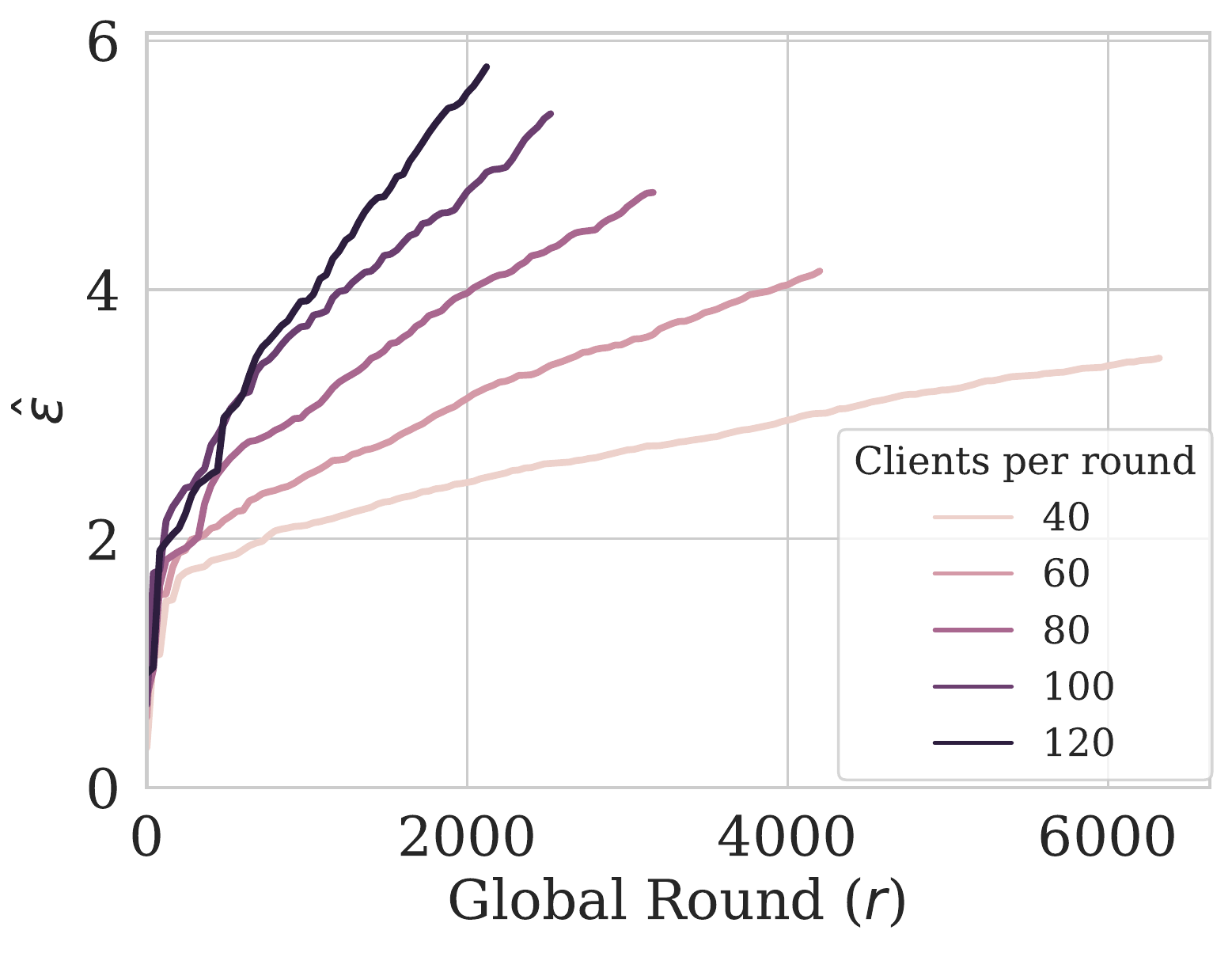}}
      \\
\caption{Further Ablations on CelebA; (a) CNN (b), (c); ResNet18}
\label{fig:appendix:ablation}
\end{figure*}

\section{Measuring Empirical Privacy} \label{appendix:empirical_priv}
\subsection{Privacy Measures and Accounting}
\edit{
Here we provide further details about privacy accounting and the different privacy measures that we analyse. In all experiments we use R\'enyi Differential Privacy (RDP) accounting with subsampling (Poisson sampling over the set of clients) to guarantee user-level DP. This is based on the \textsc{DP-FedAvg} algorithm \citep{mcmahan2017learning}  with accounting implemented via the Opacus library \citep{yousefpour2021opacus}. More specifically, the accounting uses the RDP subsampling analysis derived from \citet{mironov2019r} and the RDP to $(\eps,\delta)$-DP conversion from \citet{balle2020hypothesis}.

In this work, we have exactly four different privacy measures. For the theoretical quantities we have the per-round guarantee $\eps_r$ and final privacy guarantee $\eps$ which are computed as follows:}
\begin{itemize}
    \item \edit{$\eps_r$ - This is the per-round theoretical epsilon that is derived from the RDP accountant when the subsampling rate $q$ is set to $1$ and number of steps $R=1$. This is a constant value (dependent on the noise multiplier $\sigma$ and $\delta$) and is the privacy guarantee of performing a single step of \textsc{DP-FedAvg}.} 
    \item \edit{$\eps$ - This is the theoretical $(\eps,\delta)$-DP guarantee of the model trained under (user-level) DP. We calculate this via an RDP accountant with subsampling where the sampling rate $q$ is chosen to be the number of participating clients over the total number of clients in the population. Since we are using Gaussian noise this corresponds to the RDP analysis of the Subsampled Gaussian Mechanism (SGM), see \citep{mironov2019r} for technical details.}
\end{itemize}

\edit{Similarly, we have analogous empirical measures $\hat\eps_r$ and $\hat\eps$ which are computed as follows:}
\begin{itemize}
    \item \edit{$\hat\epsilon_r$ - Since the attack used by \method infers membership of the canary update at a particular round, the privacy measure derived from a set of \method attack scores is a per-round measure. This is computed via the formula derived in \cite{kairouz2015composition} i.e.,}
    \edit{
    \begin{align*}
      \hat{\eps}_r = \max_\gamma\left(\log \frac{1-\delta-\mathrm{FPR}_\gamma}{\mathrm{FNR}_\gamma},\log \frac{1-\delta-\mathrm{FNR}_\gamma}{\mathrm{FPR}_\gamma}\right),
    \end{align*}}\edit{where $\mathrm{FPR}, \mathrm{FNR}$ are computed from the attack scores at round $r$.} \fedit{In our experiments we maximise $\hat\eps_r$ over the threshold $\gamma$ to provide a worst-case measure.}
    \edit{The quantity $\hat\eps_r$ is directly comparable to the theoretical $\eps_r$. We also compute $95\%$ confidence intervals (CIs) for $\hat\eps_r$ from the attack scores via the Clopper-Pearson method as in \citet{nasr2021adversary}.} 
    \item \edit{$\hat\eps$ - This is the empirical privacy measure of the model derived from \method. One could apply basic composition to $\hat\eps_r$ over $R$ rounds to obtain the empirical measure of $R\hat\eps_r$ but this results in suboptimal composition. Instead we compute $\hat\eps$ under the tighter composition of RDP with amplification by subsampling. To do so, we convert each $\hat\eps_r$ into an equivalent noise multiplier $\hat\sigma_r$ and compound the noise over a number of rounds with the accountant. The quantity $\hat\eps$ is directly comparable to $\eps$. See Section \ref{appendix:alg} for more information.}
\end{itemize}

\subsection{Algorithm for $\hat\eps$}\label{appendix:alg}

In Section \ref{sec:privacy_measure}, we explained how we obtain a per-round privacy measurement $\hat\eps_r$ from \method and compound this to form a global privacy estimate $\hat\eps$ over a training run. We detail this method in Algorithm \ref{alg:measure}. In order to compound our per-round estimates $\hat{\eps}_r$ from \method, since we only attack the model every $s$ rounds we assume that the noise estimate $\hat{\sigma}_r$ remains constant between rounds $r$ and $r+s$ before we attack the model again and re-estimate the noise $\hat\sigma_r$. 

\subsection{Details for monitoring privacy}\label{appendix:monitor_details}

In Section \ref{sec:exp:monitor}, we present experiments using \method to measure empirical privacy during the training run of federated models. We ran each training run five times and examples of these runs are displayed in Figure \ref{fig:monitor}. Here we provide extra details of the training setup:
\begin{itemize}
    \item \textbf{CelebA.} We train a ResNet18 model for $30$ epochs and have 100 clients participate per training round. This results in $85$ rounds per epoch. We freeze and attack the model every $s=40$ rounds resulting in $64$ empirical privacy estimates ($\hat{\eps}_r$) across training.
    \item \textbf{Sent140.} We train for $15$ epochs and have 100 clients participate per training round. This results in $593$ rounds per epoch and $8895$ rounds in total. We freeze and attack the model every $s=100$ resulting in $90$ empirical privacy measurements across training.
    \item \textbf{Shakespeare.} We train for $30$ epochs and have $60$ clients participate per training round. This results in $17$ rounds per epoch and $510$ training rounds in total. We freeze and attack the model every $s=8$ rounds resulting in $64$ empirical privacy estimates across training.
\end{itemize}

\begin{table}[t]
    \centering
    \begin{tabular}{c} 
    \toprule 
    Canary Sample after optimization \\
    \midrule
    mRnt,,,,,,,,,,,'d,,,,,,R,,,,,A,,,,,,,,,,,,,A,,,,,,,,,,,,,,,,,AA,,,,V2,E $>$H \&3 4i\\
    Yet I confess that often ere thisMday,x?h\textbackslash nnPImh!v7DhNAd\}I!H';kXXI'PmP1Iert 6Fa \\ 
    th, nothing bu; aZ empty box, sir, wQich in my lord's beealf I cote to e? rtat 3 \\
    \bottomrule
    \end{tabular}
    \caption{Canary Samples on Shakespeare.}
    \label{tab:canary_samples}
\end{table}

\subsection{Relationship between $\hat\eps_r$ and attack accuracy}
We display the relationship between the per-round measurement $\hat\eps_r$ and the accuracy of the attack for models with a final privacy of $\eps = 50$ in Figure \ref{fig:appendix:acc}. We also found a consistent relationship for $\eps \in \{10, 30\}$.
\subsection{Further Experiments: Shakespeare}\label{appendix:empirical_priv:shakes}
In Figure \ref{fig:monitor}, we displayed example per-round estimates $\hat\eps_r$ across training runs for CelebA and Sent140. In Figure \ref{fig:appendix:shakes:eps50} we display a similar plot but for an example training run on Shakespeare. We also note that one can use \method to measure $\hat\eps_r$ for models that are trained without privacy. \fedit{In Figure \ref{fig:appendix:shakes:eps0}, we show an example training run on Shakespeare without DP ($\eps = \infty$). As mentioned in Section \ref{sec:privacy_measure}, the number of attack trials determines an upper bound on $\hat \eps_r$. Here we can see that the empirical privacy measure is essentially constant throughout (non-private) training and consistently reaches these bounds. We note that $\hat\eps_r = \infty$ only once (when the maximum attack accuracy reaches $100\%$).}

\begin{figure*}[t]
    \centering
  \subfloat[Attack accuracy \label{fig:appendix:norm:acc}]{%
      \includegraphics[width=0.4\textwidth]{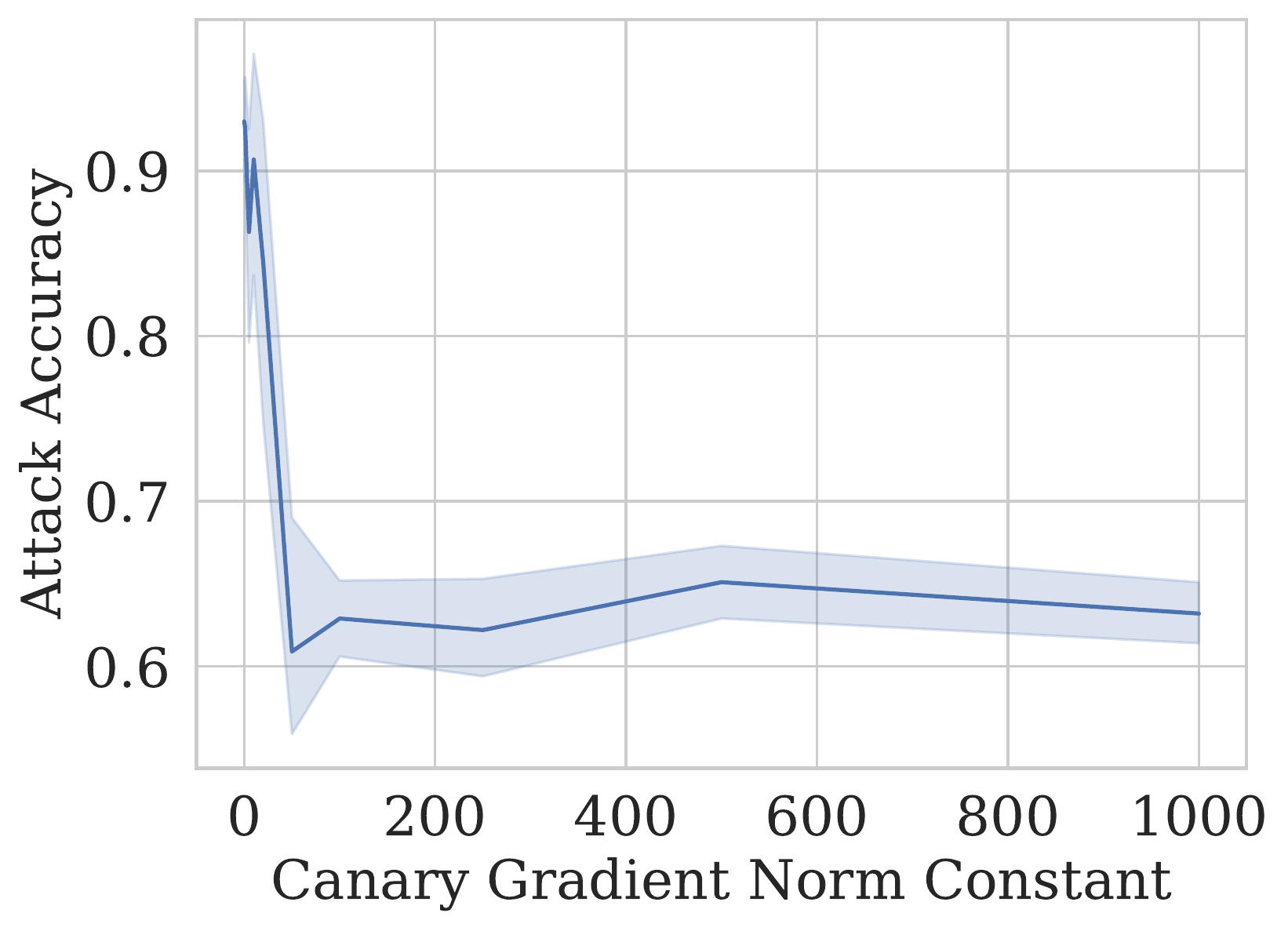}}
  \subfloat[Canary health \label{fig:appendix:norm:health}]{%
      \includegraphics[width=0.4\textwidth]{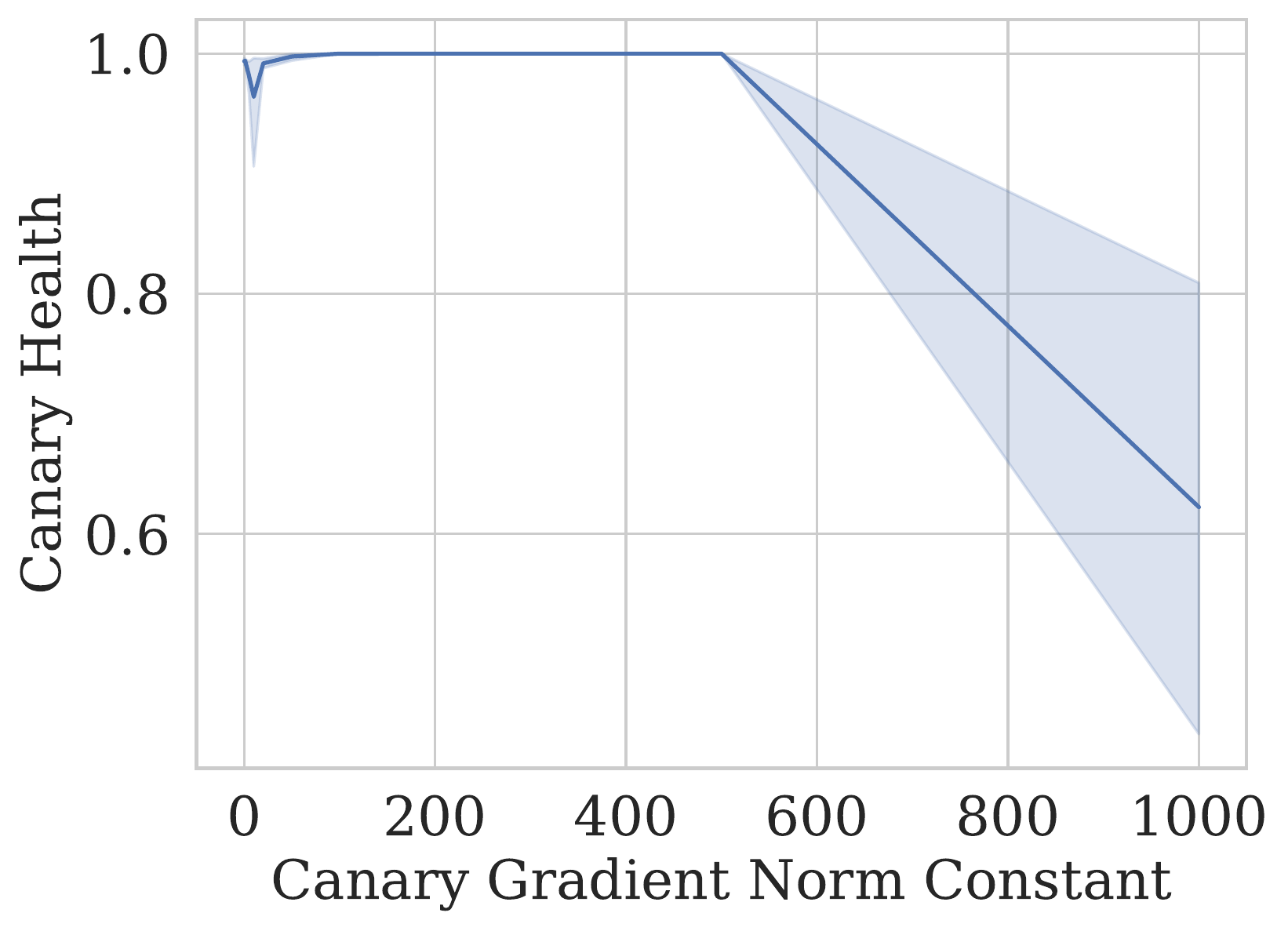}}
    \\
\caption{Varying the canary gradient norm constant on CelebA (CNN)}
\label{fig:appendix:norm}
\end{figure*}

\section{Further Ablation Studies}\label{appendix:ablation}

\paragraph{Loss modification.} As discussed in Appendix \ref{appendix:lkelihood}, we can minimise the dot-product of the canary with the average model update (\say{Loss 1}) or the covariance loss (\say{Loss 2}) which we choose to use in all experiments. In Figure \ref{fig:appendix:ablation:loss}, we vary the design sample size and carry out 5 attacks with a checkpointed CNN model on CelebA while varying these losses. We observe that there is a clear difference in accuracy between the two losses, with Loss 2 having consistently high attack accuracy which is almost constant as we increase the sample size. We note that Loss 1 seems more sensitive to the total number of design samples, with increasing attack accuracy as the design sample size increases.

\paragraph{Number of clients per round.} In Figure \ref{fig:appendix:ablation:client1}, we vary the number of clients per round and plot the average attack accuracy over 5 training runs. We note that the attack accuracy decreases slowly as the number of clients increases which is consistent with the observations made in Figure \ref{fig:cifar10:pool}. \fedit{In Figure \ref{fig:appendix:ablation:client2}, we plot the empirical estimate $\hat{\eps}$ during training. We find that although the average accuracy decreases, this does not have a significant effect on the final estimate $\hat{\eps}$ which decreases as the number of clients per round decreases.}

\paragraph{Canary gradient norm.} To conclude, we investigate the effect of the canary gradient norm constant in the second term of the loss $\mathcal{L}(z)$. We fix the privacy clipping constant to be $C=1$ and vary the gradient norm constant on CelebA. We use a checkpointed CNN model trained without DP to $70\%$ test accuracy. We attack the model $10$ times and average the results. In Figure \ref{fig:appendix:norm:acc}, we display the average accuracy of the attack and in Figure \ref{fig:appendix:norm:health} the average canary health as we vary the gradient norm constant. We observe that choosing the constant too large can significantly decrease the efficacy of the attack and that choosing the constant around $C$ is enough to guarantee high accuracy and well-behaved optimization. We note that the canary health does not significantly decrease until the norm constant is chosen to be very large ($> 600$), yet the attack suffers a large drop in accuracy for constants $> 50$. This implies that as you increase the constant, it is first possible to design a canary that has large gradient norm but at the expense of orthogonality to the other model updates, and as the constant gets very large, it becomes too difficult to optimise either term of $\mathcal{L}(z)$.

\section{Canary Samples}
\label{appendix:canary_samples}

We display in Table~\ref{tab:canary_samples} some examples of designed canaries with \method when initializing the crafting procedure with training sentences on Shakespeare.

\section{Limitations and Extensions}\label{appendix:limitations}

\edit{
In our experiments we have made assumptions for \method that may be limiting in specific practical FL scenarios. In this section, we highlight such limitations with various extensions for \method.

\paragraph{Multiple Local Epochs.} Throughout Section \ref{sec:exp}, we have restricted clients to performing a single local epoch for simplicity. But in practice clients often perform multiple local epochs and this has shown to help improve the convergence of models trained in FL. \method is not limited to methods with a single local epoch and in practice to audit methods like \textsc{DP-FedAvg} with multiple local epochs there are two main solutions:

\begin{enumerate}
    \item \textbf{Use \method as is:} There is nothing in the current formulation of \method that does not apply to multiple local updates. Since in practice, the server will mock the adversary, they are free to choose how many local epochs the canary client performs. Thus, they can design a canary with a single local epoch to be orthogonal to clients who do multiple local epochs. We also believe this is most practical for an adversary since it is the easiest from the attackers viewpoint to design and optimize for. 
    \item \textbf{Modify \method loss:} An alternative approach is to design a canary sample that has a model update (formed from multiple local epochs) that is orthogonal to all other model updates (which are also formed from multiple local epochs). To carry out such a design we can modify the \method loss in \eqref{eq:loss2} to include the canary model update $u_c$ instead of the canary gradient $\nabla_\theta \ell(z)$. In order to calculate the gradient of such a loss it requires backpropagating through the multiple local updates and essentially \say{unrolling} the local SGD steps. This may be computationally burdensome, and so while it is possible to design such a canary sample, it may not be practically viable for any adversary (and/or server) depending on the model size and number of local epochs. Thus in practice, it may be simpler to audit the model via (1). 
\end{enumerate}
}

\edit{
\paragraph{Multiple Privacy Measurements.}

One current limitation of \method is that it only produces a single privacy measurement ($\hat\eps_r$) at a specific round. In practice, model auditors may want a more comprehensive empirical analysis of the model's privacy via multiple measurements, such as attacks that vary the threat model like that of \citet{nasr2021adversary}. We believe \method can be extended to support a “multiple measurement” approach with ease since there is some degree of freedom in the canary design. For example, by using different design pools, each one strengthening the adversary further (e.g. with more data and/or design pools that better approximate the federated distribution based on prior knowledge) and thus obtain a more holistic measure of the model’s privacy. One can similarly vary the number of design epochs to simulate clients with limited computation. Designing multiple canaries per round under different constraints will generate a set of empirical epsilons that allow for more fine-grained statistics (for example, taking the maximum of the per-round empirical epsilons for a worst-case measure). 
}

\edit{
\paragraph{Preventing Wasted FL Rounds.} 
Another limitation of \method is that in our experimental setup we freeze the model for a set number of rounds to compute attack scores of a designed canary. In practice, the server and clients would be unwilling to waste federated rounds to compute \method scores. We believe \method can be extended to support a more practical attack without wasting training rounds as follows:
\begin{itemize}
    \item \textbf{Multiple Canaries:} One alternative is that the server could design multiple canaries at a single round, and use these to obtain (multiple) attack scores (subject to compute limitations). This can help reduce the number of mock rounds being run.

    \item \textbf{Running measurements:} The previous approach may be computationally prohibitive depending on server resources and still requires frozen rounds. An alternative could be to maintain “running” attack measurements where we allow \method to run alongside normal model training, letting the model change at each round. In this setup, the server can design a canary at each round, calculate attack scores and then proceed with updating the global model (without the canary inserted). This has minimal overhead to the server (who just needs to design the canary) and no additional overhead to clients (who just believe they are participating in a standard FL training round). The set of attack scores can be used to calculate empirical epsilons over various periods of training and one can change this period however they like. We emphasize that since our results (specifically Figure \ref{fig:monitor}) show the per-round empirical measure is fairly stable after the first epoch or so of training, then the approach described here should still give stable results (even though the model would be changing at each round).
\end{itemize}

}

\edit{
\paragraph{Design Pool Assumptions.} The design pool is an important component of our attack, and in order to present a conservative (worst-case) privacy measure in this work, we assume that the adversary has access to held-out data that approximates the federated training data well. To do this, we form the design pool from the test set of our datasets but in many practical scenarios this is not possible since the server may not know much about the (private) federated data. While this is a limitation of \method, we believe that for many tasks it would be possible to form a design pool from public data. For example, if the adversary knows the exact task of the model (e.g. sentiment analysis) then the adversary can form a design pool from public datasets (e.g. Sent140) or even craft their own language data for canary design. We expect that this would be a reasonable proxy for the true federated dataset and would not significantly affect privacy measurement. In scenarios where the server has no prior knowledge, they could utilise private synthetic data generation methods or Federated Analytics (FA) to privately compute statistics about client data to guide the choice of design pool. 
}

\end{document}